\documentclass[11pt,a4paper]{article}

\usepackage{booktabs} %
\usepackage{amsmath}
\usepackage{amssymb} %
\usepackage{subcaption}  %
\usepackage{hyperref}
\usepackage{cleveref}

\usepackage{enumitem}
\usepackage[lf]{sourcesanspro}
\usepackage{float}
\newcommand{\Bx}{\boldsymbol{x}}
\usepackage{threeparttable}
\usepackage{siunitx}
\newcommand{\Bq}{\boldsymbol{q}}
\newcommand{\Bk}{\boldsymbol{k}}
\renewcommand{\leq}{\leqslant}

\usepackage{orcidlink}

\usepackage[nocustomfonts]{nxai}

\newcommand{\affmark}[1]{\raisebox{0.55ex}{\fontsize{7}{7}\selectfont #1}}

\nxaisetalogo{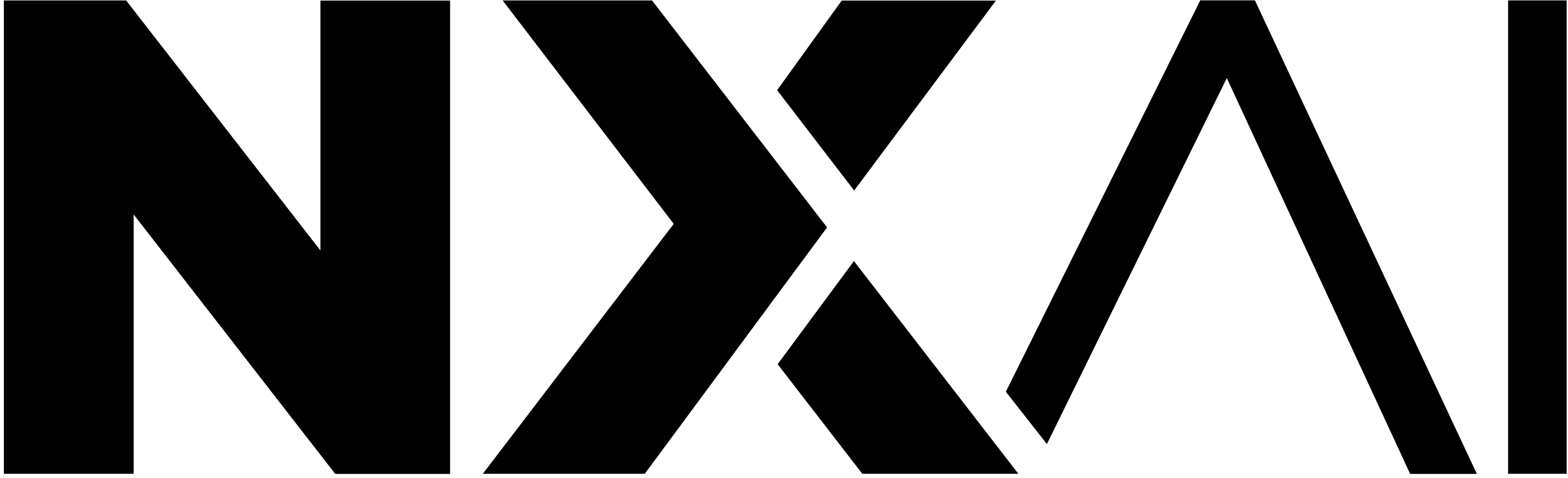} %
\nxaisetlogoheight{4mm} %
\nxaisetdate{November 23, 2026}

\nxaisetfooter{Accepted at the NeurIPS 2026 Workshop TS-LIMITS.\\
Corresponding authors: Marco Pichler, \href{mailto:pichler@ml.jku.at}{pichler@ml.jku.at}, Patrick Podest, \href{mailto:podest@ml.jku.at}{podest@ml.jku.at}}

\title{Time Series Foundation Models on the Edge}
\author{
Marco Pichler\affmark{*,1,2,3}\orcidlink{0009-0003-4425-6956}\and 
Patrick Podest\affmark{*,1,2,3}\orcidlink{0009-0003-6546-5706}\and Günter Klambauer\affmark{1,2,3}\orcidlink{0000-0003-2861-5552}\and  
Sepp Hochreiter\affmark{1,2,3}\orcidlink{0000-0001-7449-2528}
}
\nxaisetaffiliations{%
\affmark{1}ELLIS Unit Linz,  LIT AI Lab \& Institute for Machine Learning, JKU Linz, Austria, \quad
\affmark{2}NXAI Lab, Linz, Austria, \newline
\affmark{3}NXAI GmbH, Linz, Austria, \quad
\affmark{*}Equal contribution.
}

\nxaiabstract{%
  Time series foundation models (TSFMs) forecast zero-shot with accuracy exceeding that of task-specific models, and the
  smaller ones fit on edge devices. Operational deployments must refresh their forecast as new observations arrive, and
  privacy and latency requirements often force this refresh onto resource-constrained edge devices. To serve as a
  forecasting system there, a TSFM must refresh quickly, keep latency and peak memory bounded as the stream grows, and
  retain its accuracy as the observed history accumulates. Refreshing incrementally without altering the forecast
  requires temporal causality: non-causal models recompute at every update, causal attention admits a KV cache that
  grows linearly, and causal recurrent models update a fixed-size state in constant time and memory. We make these states persist across refreshes for every causal architecture and verify equivalence to full recomputation. We
  evaluate nine TSFM configurations on univariate forecasting against all three requirements, measuring accuracy as a function of context length,
  single-refresh latency, and latency and memory on streams of up to one million time steps on a server-grade GPU and three
  edge devices. No model gains on average accuracy beyond its trained context and most lose it well before the full history is reached, while KV caches leave latency and memory growing linearly with the retained history. TiRex-2 is the only one of the nine configurations meeting all three requirements: it refreshes in 14\,ms on the GPU and 120\,ms on a Raspberry Pi~5, keeps latency, energy and peak
  memory constant over a million steps, and retains its accuracy on the untruncated history.
}

\begin{document}
\maketitle

\section{Introduction}

A forecast used in operations is rarely computed once. In production-cloud
monitoring, manufacturing, intensive care, and hydrology observations arrive as
ongoing, continuously updated streams
\citep{adams2020monarch,ye_real-time_2019,zurcher2022data,lurry2010does}.
Forecasting systems must incorporate these observations quickly enough to preserve the lead time required for action
\citep{tomavsev2019clinically}.
Privacy constraints and tight response deadlines can make cloud offloading unsuitable, motivating inference near the
data source \citep{sanzmarco2020adaptive,lee2019ondevice}. Such deployments shift inference onto embedded or edge
hardware with tight compute and memory budgets \citep{sanzmarco2020adaptive}.

\begin{figure}[t]
\centering
\includegraphics{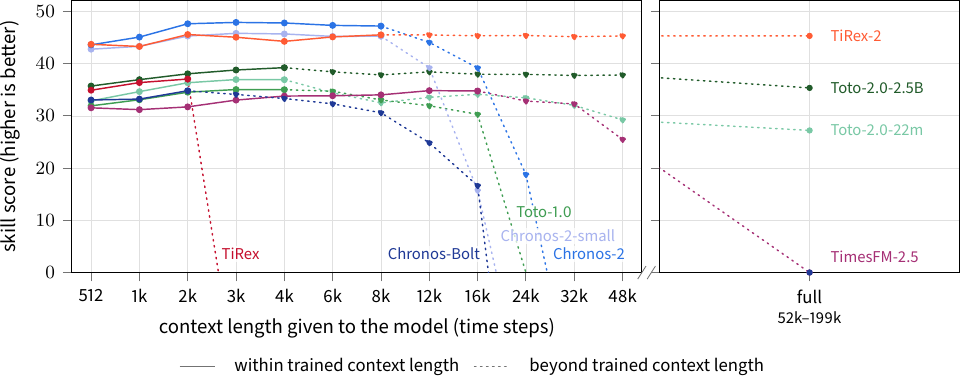}
\caption{fev-bench skill score against the context handed to the model,
aggregated over the 10 tasks whose series exceed $50\,000$ steps. Colour
identifies the architecture and each line is solid up to the model's trained
context length and dotted beyond it. No model gains on aggregate from more
context than it was trained on. Beyond it TiRex collapses, the Chronos configurations fall
below the naive baseline (off scale), the Toto family and TimesFM-2.5 degrade
gradually, and only TiRex-2 stays flat out to the untruncated series. The
rightmost rung aggregates all series ends above $48\,000$ steps ($52\,392$ to
$198\,504$ steps depending on the task).}
\label{fig:context-accuracy}
\end{figure}

Time series foundation models (TSFMs) have achieved strong zero-shot performance
\citep{ansari_chronos-2_2025,auer_tirex_2025,cohen_this_2025,podest2026tirex,khwaja2026toto20timeseries}, exceeding the accuracy of task-specific
models \citep{shchur_fev-bench_2025,aksu_gift-eval_2024}. Some TSFMs are sufficiently compact for
deployment on constrained hardware, and we therefore ask whether they can serve as forecasting systems in an
operational setting. To qualify as such a system, we require a TSFM to retain its accuracy as the observed
history accumulates, to refresh its forecast fast enough to preserve the lead time required for action, and to keep
latency and peak memory bounded as the stream grows.

As operational setting, we consider a time series whose observations arrive continuously in chunks $\Bx_n$ of $p$ time steps.
After receiving $\Bx_n$, a forecasting system $f$ updates its retained state and predicts the subsequent chunk:
\begin{align}
 (S_n, \hat{\Bx}_{n+1}) = \operatorname{Update}_f(S_{n-1}, \Bx_{n}).\label{eq:update}
\end{align}
Here, $S_n$ contains all information retained after processing chunk $\Bx_n$ with $S_0$ representing the initial state of the forecasting system.
Depending on the architecture, $S_n$ may comprise the complete observation history, a KV cache \citep{pope2023efficiently}, or a fixed-size recurrent state. We evaluate the latency and peak memory required for this update as the observed sequence grows.

Existing TSFM benchmarks primarily evaluate accuracy on fixed, predefined contexts and forecast window lengths \citep{shchur_fev-bench_2025,aksu_gift-eval_2024,Meyer_2026,qiao2026s}. fev-bench additionally reports median inference time per $100$ evaluation windows \citep{shchur_fev-bench_2025}, which does not characterize the latency or memory of refreshing a forecast after new observations arrive. To our knowledge, no prior work measures forecast-refresh latency and peak memory for TSFMs under repeated updates, or compares several of them on edge hardware (App.~\ref{sec:related-work}).

\begin{nxaiinfo}[Our contributions are as follows:]

\begin{itemize}
 \item We define a streaming evaluation protocol that measures forecast-refresh latency and peak memory as the
  observed sequence grows, while controlling for patch granularity, context truncation, batch size, and hardware, and
  apply it to nine state-of-the-art TSFM configurations on streams of up to one million time steps on an NVIDIA A40 and
  three CPU-only edge devices, a Raspberry Pi~5 and two Beckhoff industrial controllers (C6015, C6025), see
  Sec.~\ref{sec:setup}. The results distinguish three scaling regimes: full recomputation, context-growing KV caches, and
  fixed-size recurrent states.

 \item We measure the accuracy of each configuration as a function of the context length
  (\Cref{fig:context-accuracy}). No model improves beyond the context length it was trained on, and only one keeps its
  accuracy on the untruncated history. The usable history is therefore bounded by accuracy, not only by latency and
  memory.

 \item We characterize the incremental update admitted by each architecture. For every causal architecture, we carry its state across refreshes, by KV caching or recurrent state, and verify equivalence to full recomputation. For non-causal architectures, we show that recomputation is unavoidable.
\end{itemize}
\end{nxaiinfo}

\section{Methods: Model selection and streaming updates}
\label{sec:model-comparison}

We select architecturally diverse models from the leaderboard frontier, define the exactness requirement for the
update in \Cref{eq:update}, derive the latency and memory scaling of each architecture class, and modify the execution
paths to update at the lowest asymptotic complexity each architecture admits.

\begin{table}[h]
\centering
\small
\setlength{\tabcolsep}{4pt}
\caption{The nine architectures benchmarked. Patch and Context from the released configs;
complexities are those implied for one forecast over $n$ steps, measured in
App.~\ref{app:complexity}.
$^{\ast}$horizon inside a 16384-token limit.
$^{\dagger}$second ceiling at 262144 steps, rotary tables out of bounds.
$^{\ddagger}$bfloat16 as released, except scaling and aggregation; all others float32.
$^{\S}$architecture retains $\mathcal{O}(n)$; the released T5 stack materialises the
score matrix, so the measured peak is $\mathcal{O}(n^2)$.}
\label{tab:model-comparison}
\begin{tabular}{@{}llrrrrll@{}}
\toprule
Architecture & Backbone & Patch & Context & Params & Weights & Recompute & Cached \\
             &          &       &        & (M)    & (GB)    & (time\,/\,memory) & (time\,/\,memory) \\
\midrule
\multicolumn{8}{@{}l}{\emph{Causal, recurrent: fixed-size state}}\\
TiRex-2            & xLSTM & 32 &  8192             &   82.5 & 0.32 & $\mathcal{O}(n)/\mathcal{O}(n)$   & $\mathcal{O}(1)/\mathcal{O}(1)$ \\
TiRex              & xLSTM & 32 &  2048             &   35   & 0.13 & $\mathcal{O}(n)/\mathcal{O}(n)$   & $\mathcal{O}(1)/\mathcal{O}(1)$ \\
\midrule
\multicolumn{8}{@{}l}{\emph{Causal, attention: state grows with context}}\\
Toto-1.0           & decoder-only     & 64 &  4096             &  151   & 0.57 & $\mathcal{O}(n^2)/\mathcal{O}(n)$ & $\mathcal{O}(n)/\mathcal{O}(n)$ \\
Toto-2.0-22m       & decoder-only     & 32 &  4096$^{\dagger}$ &   22   & 0.10 & $\mathcal{O}(n^2)/\mathcal{O}(n)$ & $\mathcal{O}(n)/\mathcal{O}(n)$ \\
Toto-2.0-2.5B      & decoder-only     & 32 &  4096$^{\dagger}$ & 2500   & 9.36 & $\mathcal{O}(n^2)/\mathcal{O}(n)$ & $\mathcal{O}(n)/\mathcal{O}(n)$ \\
TimesFM-2.5        & decoder-only     & 32 & 16256$^{\ast}$    &  200   & 0.86 & $\mathcal{O}(n^2)/\mathcal{O}(n)$ & $\mathcal{O}(n)/\mathcal{O}(n)$ \\
\midrule
\multicolumn{8}{@{}l}{\emph{Non-causal: no incremental state}}\\
Chronos-2         & T5 encoder & 16 & 8192 &  120 & 0.45              & $\mathcal{O}(n^2)/\mathcal{O}(n)$      & --- \\
Chronos-2-small   & T5 encoder & 16 & 8192 &   28 & 0.10              & $\mathcal{O}(n^2)/\mathcal{O}(n)$      & --- \\
Chronos-Bolt-base & T5 encoder & 16 & 2048 &  205 & 0.38$^{\ddagger}$ & $\mathcal{O}(n^2)/\mathcal{O}(n)^{\S}$ & --- \\
\bottomrule
\end{tabular}
\end{table}
\subsection{Model selection}
\label{sec:selection}

We use a reproducible selection rule based on the fev-bench leaderboard
\citep{shchur_fev-bench_2025}. Its Pareto frontier of skill score and median
inference time contains four configurations from three model families: TiRex
\citep{auer_tirex_2025}, TiRex-2 \citep{podest2026tirex}, Toto-2.0-22m
\citep{khwaja2026toto20timeseries}, and Chronos-2
\citep{ansari_chronos-2_2025} (\Cref{fig:pareto}). To separate
architectural effects from model generation and scale, we additionally include
the previous generation and, where available, the smallest and largest
checkpoint of each represented family. This adds Toto-1.0
\citep{cohen_this_2025}, Toto-2.0-2.5B
\citep{khwaja2026toto20timeseries}, Chronos-Bolt-base
\citep{ansari_chronos_2024}, and Chronos-2-small
\citep{ansari_chronos-2_2025}. The two TiRex generations are already on the
frontier. Finally, we include TimesFM-2.5 \citep{das_decoder-only_2024} because
it has the largest trained context length $p\times n$ among the leaderboard entries. This procedure
yields nine configurations, see \Cref{tab:model-comparison}. The leaderboard inference times are used only for
selection and not as measurements in our evaluation.

\subsection{Operational updates and exactness}

Let $\hat{\Bx}_{n+1}^{\mathrm{inc}}$ denote the forecast produced by the incremental update (\Cref{eq:update}) and
$\hat{\Bx}_{n+1}^{\mathrm{full}} = f(\Bx_{1:n})$ the forecast of a full recomputation by $f$ on the complete observed
context. We require the two to agree up to numerical precision:
\begin{align}
  \hat{\Bx}_{n+1}^{\mathrm{inc}} = \hat{\Bx}_{n+1}^{\mathrm{full}}. \label{eq:correctness}
\end{align}
The condition lets us compare execution strategies without changing the forecast. Our conclusion does not rest on it:
granting every architecture its best measured truncation window instead leaves the ordering unchanged
(App.~\ref{app:windowing}).
This condition can always be satisfied by recomputing the model. Reusing previously computed representations, however,
requires temporal causality, i.e., new observations must not cause a change in representations computed from earlier
observations. 
This distinction allows us to classify the selected models into three architecture classes. Non-causal transformers
require full recomputation because new observations change earlier representations. Their update time therefore scales
as $\mathcal{O}(n^2)$, while retaining the complete observation history requires $\mathcal{O}(n)$ memory. Causal
transformers admit a KV cache: each update adds a constant amount to the cache, but the total retained state and the
time required to attend to it both scale as $\mathcal{O}(n)$. Causal recurrent models maintain a fixed-size state,
yielding $\mathcal{O}(1)$ update time and $\mathcal{O}(1)$ retained memory, with no growth in the retained state after
initialization.
\Cref{tab:model-comparison} collects the scaling properties of the selected models. Its \emph{Recompute} column gives
the cost of a full forward pass over $n$ steps, and its \emph{Cached} column the cost of one incremental update, which
is the update we time for every causal model, attention-based and recurrent alike.

\subsection{Model implementation changes and validation}
\label{sec:implementation}
We make the retained state of every causal architecture persist across forecast calls. Only the execution path changes.
Preprocessing, patch embedding, weights, and forecast head remain as released. The non-causal Chronos configurations
admit no reusable temporal state and recompute the observed context in full at every update. To evaluate contexts beyond the trained lengths, we remove the released context caps where they
are policy checks rather than architectural limits.

In both causal families the reusable state is recorded per layer and consumed at the next call: at every attention layer of the causal transformers the keys and values of the arriving chunk are appended to that layer's cache, and in the causal recurrent models we retain the xLSTM cell state of every block. A full pass over the initial context
initializes the state, while each subsequent update projects only the new chunk, advances the per-layer state, and
applies the forecast head to the new final representation. TiRex-2 exposes a native single-step update whose per-block
states we thread through the complete forecasting stack, whereas TiRex discards its recurrent state after each block, so
we adapt its interface to return that state and accept it.

We validate every incremental implementation against the correctness condition, \Cref{eq:correctness}. At a context
of $2048$ steps we compare the forecast obtained by ingesting the context incrementally with that of a full forward
pass over the same observations and forecast horizon. We accept an implementation if the two pass \texttt{allclose} with $\mathrm{atol}=10^{-5}$ and $\mathrm{rtol}=10^{-4}$.
Every model passes, with a maximum absolute deviation below $10^{-6}$ (\Cref{tab:equivalence}).

\section{Experiments}
We conduct three experiments to evaluate whether a TSFM meets the three requirements of operational use: accuracy
retained as the context expands, a forecast refresh that completes within the interval between arriving chunks, and
latency and memory that stay bounded as the stream grows.

\paragraph{Experimental setup.}
\label{sec:setup}
We measure on one server-grade accelerator and three edge devices. The server-grade reference is an NVIDIA
A40\footnote{NVIDIA A40 (GA102, Ampere, 48~GB GDDR6 with ECC enabled, 300~W power limit, driver 590.48.01 (CUDA 13.1)) in
a dual-socket host with 2$\times$ Intel Xeon Platinum 8168 @ 2.70~GHz (48~cores / 96~threads), 754~GB DDR4, Rocky
Linux~9.8, kernel 5.14.0-687.31.1.el9\_8, x86\_64. Single GPU runs only.}. The edge devices run inference on the CPU
only. They are a Raspberry Pi~5\footnote{Raspberry Pi~5 Model~B
Rev~1.1: Broadcom BCM2712, 4$\times$ ARM Cortex-A76 @ 2.4~GHz, 8~GB LPDDR4X, Debian~13, kernel 6.12.62-rpi-2712,
aarch64.}, a widely used single-board computer, and two Beckhoff industrial PCs, a C6015\footnote{Beckhoff C6015-0030: Intel Atom x6416RE @ 1.70~GHz (Elkhart Lake,
4~cores), 8~GB LPDDR4X, Windows~11 IoT Enterprise LTSC 10.0.26100, x86\_64.} and a C6025\footnote{Beckhoff C6025-0010: as the C6015, but with an Intel Core i5-1145G7E
@ 2.60~GHz (Tiger Lake, 4~cores, SMT disabled) and 16~GB LPDDR4X.}, which are
controller-class machines of the kind deployed next to production equipment in manufacturing. No thermal throttling
was observed on any device during the experiments.

\subsection{Experiment 1: Accuracy over context expansion}
\label{sec:accuracy-context}

Each arriving chunk grows the observed context, so we measure zero-shot accuracy against the context length given to a
model. We therefore take the ten tasks from fev-bench \citep{shchur_fev-bench_2025} with at least $50\,000$ time steps,
which exceeds every trained context length. Let $W$ denote the number of most recent time steps a model conditions on.
We evaluate each model at increasing $W$ up to the untruncated series. Each $W$ also gives the accuracy of a fixed
truncation window of that length (App.~\ref{app:windowing}). As metric we use the fev-bench skill score against a Seasonal Naive
baseline, aggregated as the geometric mean of the per-task loss ratios. The runs are deterministic and
device-independent, so no timing is reported here.

\Cref{fig:context-accuracy} shows that no model attains a better aggregate score beyond its trained context length than at or below it. Individual tasks may peak later (App.~\ref{app:accuracy}).
Beyond that point, eight of the nine architectures deteriorate with various speeds. TiRex is the fastest to fall below the
baseline (0) set by the Seasonal Naive. The attention-based architectures show different slopes of deterioration, with
the Toto family degrading most slowly among the attention-based models. A candidate explanation is the XPOS \citep{sun2023length} length extrapolation
mechanism used in the Toto models, which our measurements do not isolate. TiRex-2 sets itself apart at no loss of accuracy, in contrast to its previous
iteration TiRex. The added mLSTM blocks are the only architectural change between the two
generations that is apparent in the univariate case, and hence a candidate explanation. Trained context, size and corpus differ as well, but a longer trained context alone would move the breakdown rather than remove it.

\begin{figure}[ht]
\centering
\includegraphics{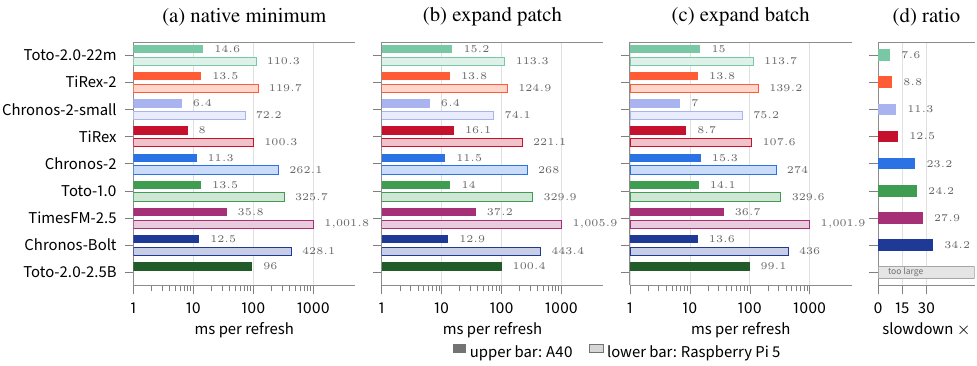}
\caption{Median latency of one refresh, base batch size 1, log scale in (a)--(c).
Upper bar per row is an A40, lower bar a Raspberry Pi 5; colour identifies the
model. Every row times the same amount of work: one forward pass, and for
Toto-1.0, the only configuration with a sampling head, one sampled path. Panel
(d) gives the Pi/A40 ratio under the native-minimum protocol; rows are ordered
by this ratio, largest at the top. The ordering stays constant across the three
protocols. App.~\ref{app:refresh-fev} plots fev-bench skill score against these
latencies.}
\label{fig:refresh-latency}
\end{figure}

\subsection{Experiment 2: Single-chunk prediction latency}
\label{sec:single-chunk}

We measure the latency of predicting one chunk from one chunk of input on the A40 and the Raspberry~Pi~5, at batch
size~1. Let $p_f \in \{16,32,64\}$ denote the native patch size of model $f$ and let $p_{\mathrm{common}}=64$ be the
shortest chunk that contains a whole number of patches for every model. We evaluate three protocols:

\begin{enumerate}[label=(\alph*), leftmargin=*, itemsep=2pt,
                  topsep=4pt, parsep=0pt]
  \item \textbf{Native chunk.}
  Each model receives $p_f$ time steps and predicts the next $p_f$ time steps. This measures the model smallest
  possible latency, but differs in the amount the model ingests and subsequently predicts.

  \item \textbf{Common chunk along the sequence axis.}
  Here $p_f=p_\mathrm{common}$ resulting in 4, 2, or 1 patches predicted by the model depending on its native patch
  size.

  \item \textbf{Common chunk along the batch axis.}
  Again $p_f=p_\mathrm{common}$, but instead of predicting multiple patches in sequence to reach $p_f$ time steps, the
  input is distributed 4, 2, or 1 times across the batch dimension. The model then again predicts $p_f$ time steps.

\end{enumerate}

\paragraph{At one chunk of context, latency is set by the call rather than by the number of patches in it.}
If latency scaled with the number of processed patches, the matched protocols (b) and (c) would cost $4\times$,
$2\times$ and $1\times$ the native latency at patch sizes 16, 32 and 64. \Cref{fig:refresh-latency} shows that they do
not: eight of the nine configurations refresh within $1.35\times$ of their native latency under both matched protocols.
At one chunk a call does little arithmetic, so its latency is dominated by a fixed per-call cost rather than by the
patches it processes.
TiRex is the only exception, and only along the sequence axis, where it emits one output patch per forward pass and
doubles its latency. Distributing the same chunk across the batch leaves its latency unchanged. Moving from the A40 to the
Raspberry~Pi~5 slows down every configuration that fits on it, but by very different factors. Toto-2.0-22m and TiRex-2
lose the least, $7.6\times$ and $8.8\times$, while Chronos-Bolt loses more than $30\times$ (panel~(d)). Toto-2.0-2.5B does not fit the $7.9$~GiB available on the Raspberry~Pi~5 at all. A single refresh of Toto-2.0-22m or
TiRex-2 therefore survives the move to edge hardware best.

\begin{figure}[h!]
\centering
\includegraphics{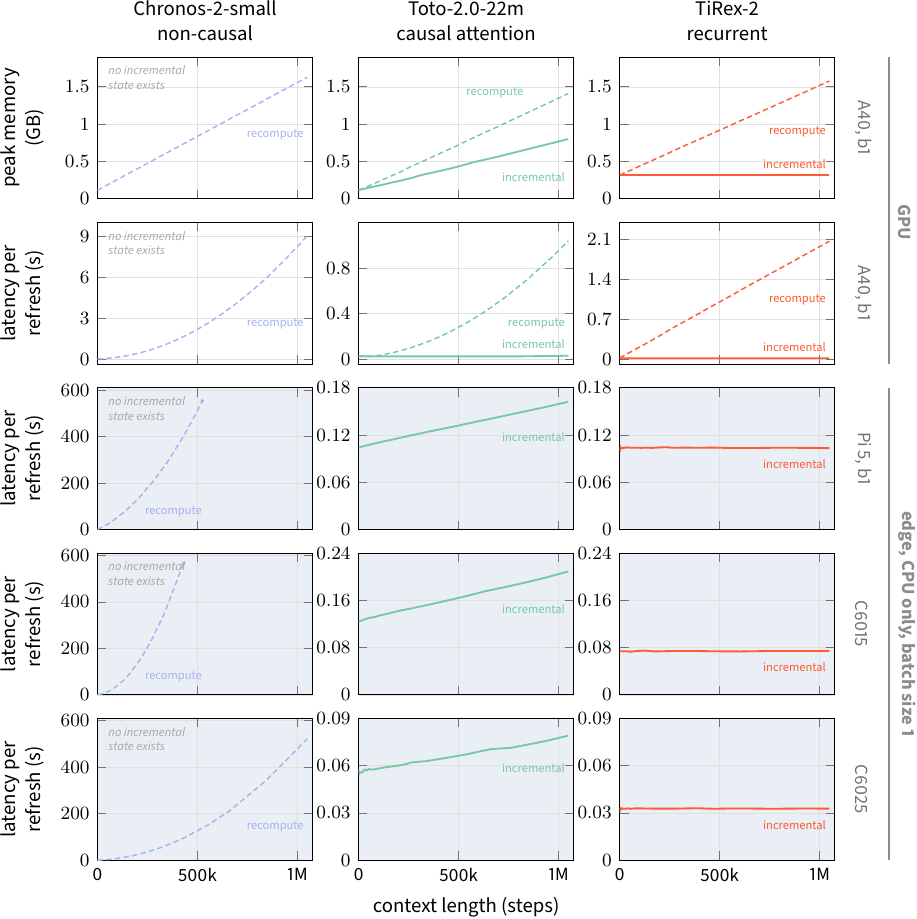}
\caption{Latency and memory scaling behavior of Chronos-2-small, Toto-2.0-22m and TiRex-2 as function of context length on an NVIDIA A40 GPU (top two rows) as well as latency behavior on a Raspberry Pi 5, a Beckhoff C6015 and a Beckhoff C6025 controller (bottom three rows).
Dashed recomputes the context, solid performs a stateful incremental update by one patch. Context lengths are multiples of $192$ steps, so that every model sees whole patches.}
\label{fig:streaming}
\end{figure}

\subsection{Experiment 3: Latency and memory scaling}
\label{sec:streaming}

\begin{figure}[t]
\centering
\includegraphics{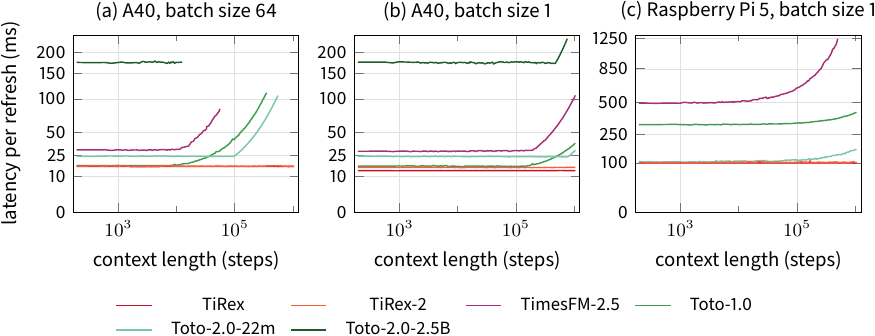}
\caption{Latency of one stateful incremental update as function of context length compared across the causal architectures. A curve ending before reaching $1$M steps indicates termination due to out-of-memory or exceedance of the per-update time budget of 600 seconds.}
\label{fig:streaming-latency-cached}
\end{figure}

As an extension of Experiment~2, we grow the number of chunks in each model's context up to $1$M steps, well past
every trained context, and measure latency and peak memory as a function of it. All configurations see the same
synthetic stream at the same context lengths, on the A40 at batch sizes 64 and 1 and on the edge devices at batch
size 1. Where both paths exist, we time full recomputation and the incremental update. A curve ends where a run
exhausts memory or exceeds a per-update budget of $600$\,s. Figure~\ref{fig:streaming} shows the fastest
configuration of each architecture class on the edge devices, Figure~\ref{fig:streaming-latency-cached} the
incremental path of every causal configuration, and App.~\ref{app:additionalresults} the remaining ones.

The three classes separate as \Cref{tab:model-comparison} predicts. Recomputation grows by three orders of magnitude
over the grid on every device, and on the edge devices a single refresh takes minutes long before $1$M steps. The KV cache keeps latency far below recomputation, but it grows linearly with the context: at batch size 64 every causal
attention configuration exhausts the A40's $48$\,GB before $1$M steps, and TimesFM-2.5 also exhausts the memory of the
Raspberry~Pi~5. Only the recurrent TiRex and TiRex-2 stay flat in both latency and memory and reach $1$M steps on every
device. TiRex-2 refreshes in $14$\,ms on the A40 and in about $120$\,ms on the Raspberry~Pi~5, from the first grid
point to the last.

\section{Conclusion}
\label{sec:conclusion}
We investigated whether leading TSFMs meet the three requirements of operational, resource-constrained forecasting.
We find that causality determines whether a model's state can be cached either in a KV cache or a compressed recurrent
memory. KV caches grow linearly with the observed context, so every causal attention-based architecture exhausts a
fixed memory budget at some stream length.
In contrast, the two fixed-size recurrent state architectures, TiRex and TiRex-2, admit constant updates in latency and
memory, with a one-chunk refresh latency within $2.2\times$ of the fastest model and about half a joule spent per
incremental update (App.~\ref{app:energy}).
Moreover, TiRex-2 is the only model that retains its accuracy well beyond its trained context, while the others
must truncate the growing history to keep theirs. When truncation is allowed, Toto-2.0-22m is the fastest model on most edge devices at a common window of $2048$
steps, but TiRex-2 scores about nine skill points higher at two to three times its latency, and its incremental refresh
on the full history is faster still (about $120$\,ms against $141$\,ms on the Raspberry~Pi~5;
Fig.~\ref{fig:context-accuracy}, App.~\ref{app:windowing}). TiRex-2 is therefore the only one of the nine
configurations to meet all three requirements.

We evaluate univariate forecasting only. Covariates scale the per-update work and the retained state by a constant
factor and leave the three regimes unchanged.
\citet{podest2026tirex} evaluate the multivariate forecasting accuracy of TiRex-2 under context extension, in a
setting similar to the operational one of this study.

\section*{Acknowledgements}
We thank Elias Bürger, Wilhelm Berghammer and Sebastian Böck for discussions of the experiments and measurements, and
Christian Ganhör for his advice on the edge devices.
The LIT AI Lab and the Institute for Machine Learning are supported by the Federal State Upper Austria.

\bibliographystyle{plainnat}
\bibliography{references}

\clearpage
\appendix
\crefalias{section}{appendix}
\crefalias{subsection}{subappendix}
\crefalias{subsubsection}{subsubappendix}
\startcontents[appendix]

\section*{Appendix}
\printcontents[appendix]{l}{1}{\setcounter{tocdepth}{2}}
\clearpage

\section{Related work}
\label{sec:related-work}

\paragraph{Efficiency evidence.}
Most TSFM papers center their evaluation on forecasting accuracy, while
inference latency and memory receive little attention
\citep{ansari_chronos_2024,ansari_chronos-2_2025,das_decoder-only_2024,liu_moirai_2026}.
Notable exceptions address different parts of the efficiency problem. Tiny Time
Mixers reports batch-32 inference time and maximum GPU memory, and Toto-1.0
reports wall-clock time, CUDA time, peak memory, and FLOPs for several TSFMs
\citep{ekambaram_tiny_2024,cohen_this_2025}. Both use fixed windows and
server-class hardware rather than operational forecast refreshes. APEX instead
evaluates its 10.5M-parameter model on a Raspberry Pi~5, but does not compare
other TSFMs on that device \citep{pradhan2026apex}. TiRex-2
targets continuous operation and demonstrates stateful forecasting for streams
of up to 32M steps at constant per-patch latency, but does not report
measured per-update latency or memory \citep{podest2026tirex}. In contrast, the
present work jointly evaluates inference latency and memory for a set of TSFMs
under a common operational protocol.

\paragraph{Architectural scaling.}
Recurrent and state-space TSFMs motivate favorable scaling through different
mechanisms: TiRex and TiRex-2 use state-tracking xLSTMs
\citep{auer_tirex_2025,podest2026tirex}, FlowState uses a parallelizable
state-space encoder \citep{graf_flowstate_2025}, Reverso combines subquadratic
long convolutions with linear RNN layers \citep{fu_reverso_2026}, and TempoPFN
argues for fully parallelizable linear RNNs as a scalable alternative to
nonlinear RNNs and transformers \citep{moroshan2026tempopfn}. These claims are
architectural or asymptotic and are not a controlled comparison of measured update latency and memory. For attention-based TSFMs, causality determines the available incremental
path: under exact full-history semantics, bidirectional models such as Chronos-2 must recompute representations when
the context changes \citep{ansari_chronos-2_2025}, whereas causal models can
reuse a KV cache \citep{pope2023efficiently}, reducing update computation while
retaining state that grows with the processed history. A recurrent model can in
principle update a fixed-size state instead. We therefore compare full
recomputation, cached attention, and recurrent state rather than treating all
architectures as equivalent fixed-window predictors.

\paragraph{Operational deployments.}
Task-specific time-series models currently operate in production. An LSTM
ensemble provides public seven-day flood forecasts in over 80 countries
\citep{nearing_global_2024}. \citet{sprangers2024hierarchical} propose a
hierarchical forecasting method used daily in production, while
\citet{ren2019anomaly} propose a streaming anomaly detector deployed
across over four million series for more than 200 teams. These systems run
repeatedly as data arrive, making per-update latency and memory practical
concerns. We therefore evaluate TSFMs under both constraints.

\section{Pareto frontier}
\label{app:selection}
\begin{figure}[!htbp]
\centering
\includegraphics{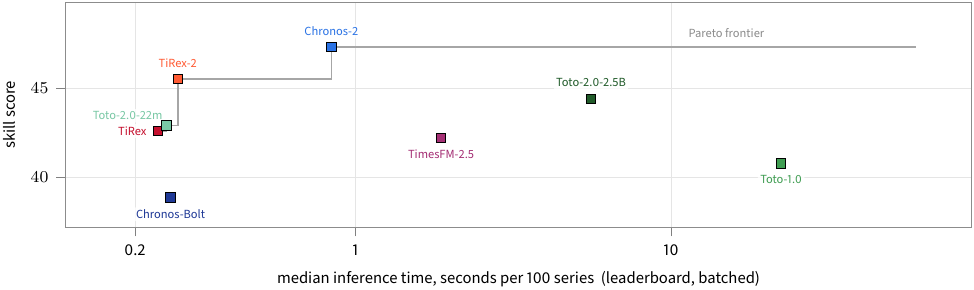}
\caption{The models this paper benchmarks, as they appear on the fev-bench
leaderboard: skill score against median inference time, with the Pareto
frontier drawn as a staircase. The frontier is the one computed over all
fifteen zero-shot entries of the leaderboard but only the selected models (see Sec.~\ref{sec:selection}) are shown, except for Chronos-2-small which the leaderboard does not
list.}
\label{fig:pareto}
\end{figure}

\newpage

\newpage
\section{Accuracy against refresh latency}
\label{app:refresh-fev}
\begin{figure}[!htbp]
\centering
\includegraphics{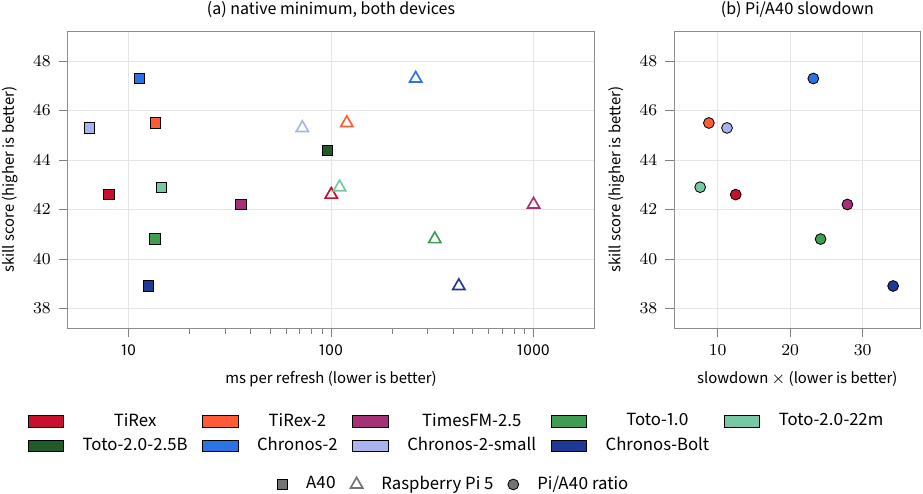}
\caption{Skill score on fev-bench against single-refresh latency, base batch
size 1, native-minimum protocol. In (a) each architecture appears twice at the
same height and in the same colour: filled square for the A40, open triangle for
the Raspberry Pi~5. The horizontal gap between the two is the slowdown that (b)
plots directly. As in Fig.~\ref{fig:refresh-latency}, every point times the same
amount of work, one sampled path for Toto-1.0 and one forward pass otherwise.}
\label{fig:refresh-latency-fev}\end{figure}

The rest of this paper measures latency and memory alone. Whether a fast refresh
also forecasts accurately is a separate question, and it decides how the main
results should be read: if the fastest architectures were also the least
accurate ones, the three regimes of Sec.~2 would describe a trade-off and every
latency number would carry an implied accuracy penalty. We therefore place all
nine configurations on both axes at once, fev-bench skill score against the
latency of one refresh, on the A40 and on the Raspberry~Pi~5
(Fig.~\ref{fig:refresh-latency-fev}). Skill is better when larger and latency
when smaller, so the upper left corner of either panel is the desirable one.

The two quantities are unrelated, and where an ordering is visible it runs the
wrong way for a trade-off. The fastest architecture on the A40 is
Chronos-2-small at $6.4$~ms, which is also the third most accurate of the nine;
the slowest is Toto-2.0-2.5B at $96$~ms, $15\times$ behind and only $0.9$ points
lower in skill. Between those two ends the ordering by latency reproduces
neither on the Pi nor in the slowdown: the remaining seven configurations sit
within a factor of $4.5$ of each other on the A40 while their skill scores span
$8.4$ points. Refresh latency is therefore not traded against skill, and the
regimes may be read without an accuracy footnote.

Panel (b) resolves the same point by device. Toto-2.0-22m degrades least from
A40 to CPU at $7.6\times$ and scores $42.9$, TiRex-2 sits at $8.8\times$ and
$45.5$, Chronos-2-small at $11.3\times$ and $45.3$, and Chronos-2 reaches the
highest skill of the set, $47.3$, at $23.2\times$. TiRex-2 scores $1.8$ points below Chronos-2 while degrading $2.6\times$ less, and gains $2.6$ points over
Toto-2.0-22m for a $1.2\times$ larger slowdown; 
Chronos-Bolt marks the other corner at $34.2\times$ and the lowest skill.
Toto-2.0-2.5B has no Pi run, as $9.36$~GB in fp32 does not fit $7.9$~GiB.

Skill scores in this appendix section are the fev-bench scores over all tasks, not the
ten-task aggregate of Tab.~\ref{tab:acc-agg}, so that the accuracy axis is not restricted to the
subset the context experiments run on. Chronos-2-small, which the leaderboard does not list, takes
the score reported by \citet{ansari_chronos-2_2025}.

\newpage

\section{Empirical verification of the asymptotic scaling}
\label{app:complexity}
We fit every result with $y = b + a\,n^{\alpha}$, profiling $\alpha$ on the last octave
of the grid, and with \newline $y = c_0 + c_1 n + c_2 n^{2}$, the form an attention stack implies.
The local exponent of the second form is $\alpha(n) = 1 + (1 + n^{\ast}/n)^{-1}$ with
$n^{\ast} = c_1/c_2$, so a quadratic architecture approaches $2$ and reaches it only once
the context is well past its own crossover. Column $\alpha_{\mathrm{pred}}$ evaluates this
at the end of each grid; it is what the theory predicts a finite measurement should show.

\begin{table}[h]
\centering
\small
\setlength{\tabcolsep}{5pt}
\caption{Complexities of Table~\ref{tab:model-comparison} against the measurement, A40 at
batch size~1, the longest grid. The other measurement cells agree with their own
$\alpha_{\mathrm{pred}}$ to within $0.2$. Incremental columns give the slope of a
floor-plus-linear fit and of the retained state; $\leq x$ is a 95\,\% upper bound. Every
entry recovers the complexity of Table~\ref{tab:model-comparison}, whose recompute memory
claim is $\mathcal{O}(n)$ throughout and $\mathcal{O}(n^2)$ for Chronos-Bolt.}
\label{tab:complexity}
\begin{tabular}{@{}lccccrr@{}}
\toprule
 & \multicolumn{4}{c}{recompute} & \multicolumn{2}{c}{incremental} \\
\cmidrule(lr){2-5}\cmidrule(l){6-7}
 & \multicolumn{3}{c}{latency} & memory & ms per & MB per \\
Architecture & claim & $\alpha_{\mathrm{pred}}$ & $\hat{\alpha}$ & $\hat{\alpha}$ & $10^{6}$ & $10^{6}$ \\
\midrule
\multicolumn{7}{@{}l}{\emph{Causal, recurrent: refresh and state independent of the stream, $\mathcal{O}(1)/\mathcal{O}(1)$}}\\
TiRex-2         & $\mathcal{O}(n)$   & 1.04       & 0.96 & 1.00 & $\leq 0.04$ & $0$ \\
TiRex           & $\mathcal{O}(n)$   & 1.05       & 0.97 & 1.00 & $0.049$     & $0$ \\
\midrule
\multicolumn{7}{@{}l}{\emph{Causal, attention: cache bounds growth to linear, $\mathcal{O}(n)/\mathcal{O}(n)$}}\\
Toto-1.0        & $\mathcal{O}(n^2)$ & 1.53       & 1.43 & 1.00 & 22.8 & 1\,007 \\
Toto-2.0-22m    & $\mathcal{O}(n^2)$ & $\to 2$    & 1.83 & 1.00 & 23.6 & 666 \\
Toto-2.0-2.5B   & $\mathcal{O}(n^2)$ & 1.80       & 1.72 & 1.04 & 231  & 23\,387 \\
TimesFM-2.5     & $\mathcal{O}(n^2)$ & 1.37       & 1.26 & 1.00 & 90.7 & 6\,108 \\
\midrule
\multicolumn{7}{@{}l}{\emph{Non-causal: no reusable state, every refresh re-encodes}}\\
Chronos-2       & $\mathcal{O}(n^2)$ & 1.91       & 1.87 & 1.00 & -- & -- \\
Chronos-2-small & $\mathcal{O}(n^2)$ & 1.94       & 1.93 & 1.00 & -- & -- \\
Chronos-Bolt    & $\mathcal{O}(n^2)$ & $\to 2$    & 2.03 & 2.00 & -- & -- \\
\bottomrule
\end{tabular}
\end{table}

Chronos-Bolt-base is the one architecture whose recompute memory is quadratic rather than
linear. It takes the eager attention of the released T5 stack, which has no fused path
because the relative position bias is added to the scores, so every layer materialises a
dense score tensor and the fp32 softmax over it holds two further copies, $12$ bytes per
score element. The peak therefore follows the score grid rather than the retained history,
which stays $\mathcal{O}(n)$: $32.9$\,GB at $248\,448$ steps, where Chronos-2 needs
$1.0$\,GB because rotary embeddings let it dispatch to a tiled kernel. Removing either
copy changes the numerics, which the exactness requirement rules out.

Two further checks support the table. In every attention and encoder row $c_2$ is
positive at $4.5$ to $115$ standard errors, while for TiRex and TiRex-2 it is compatible
with zero or implies $n^{\ast} > 1.8\!\cdot\!10^{7}$ steps. On the incremental path the
exponent above the floor is $0.91$ to $1.11$, no row is quadratic, and the recurrent peak
memory is bit-identical across all $115$ grid points.
\newpage

\section{Energy consumption}
\label{app:energy}

The Raspberry\,Pi\,5 exposes the power delivered to its supply rails. We sample it every
$0.3\,\mathrm{s}$ in separate runs that repeat one configuration for about a minute, averaging over
the timed block alone so that loading and priming are excluded; the energy of one refresh is that average times the latency in the table; for
Toto-1.0 this is the energy of one sampled path. A refresh that advances a state leaves the four cores far
from saturated and draws $4.40$ to $5.03\,\mathrm{W}$, a recompute pass $5.54$ to
$6.84\,\mathrm{W}$; the USB-C input takes $11$--$18\,\%$ more than these rail figures. The Beckhoff
controllers carry no such sensor. At the end of the grid a fixed-size-state refresh stays below one
joule, against $991\,\mathrm{J}$ to re-encode the same context.

\begin{table}[H]
\centering
\caption{Energy of one refresh on the Raspberry\,Pi\,5, batch size~1, as the power of a separate run
of the same configuration times the latency measured for it. Upper block: the native-minimum call of
Fig.~\ref{fig:refresh-latency}(a), one patch in and one forward pass out, so the last column divides
the patch out. Lower block: the end of the streaming grid, Figs.~\ref{fig:apx-edge-regimes}
and~\ref{fig:apx-chronos}. The uncertainty is the reproducibility of the power and of the latency
added in quadrature; within a session the power repeats to $1.8\,\%$, across days to $3.5\,\%$.
Sample scatter within a block is larger, $9$--$12\,\%$, but that is the \texttt{ondemand} governor
stepping the clock.}
\label{tab:energy-edge}
\begin{tabular}{@{}l r r r r@{}}
\toprule
Architecture & patch & latency & energy & energy per \\
 & & ms & J & forecast step, mJ \\
\midrule
\multicolumn{5}{@{}l}{\emph{Native minimum, one patch in and one patch out}} \\
TiRex-2           & 32 &      119.7 & $0.582 \pm 0.011$  & $18.2 \pm 0.3$ \\
TiRex             & 32 &      100.3 & $0.510 \pm 0.010$  & $15.9 \pm 0.3$ \\
Toto-1.0$^\dagger$ & 64 &     325.7 & $1.565 \pm 0.029$  & $24.4 \pm 0.5$ \\
Toto-2.0-22m      & 32 &      110.3 & $0.491 \pm 0.009$  & $15.4 \pm 0.3$ \\
TimesFM-2.5       & 32 &   1\,001.8 & $5.124 \pm 0.096$  & $160.1 \pm 3.0$ \\
Chronos-2         & 16 &      262.1 & $1.314 \pm 0.025$  & $82.1 \pm 1.5$ \\
Chronos-2-small   & 16 &       72.2 & $0.360 \pm 0.007$  & $22.5 \pm 0.4$ \\
Chronos-Bolt-base & 16 &      428.1 & $2.153 \pm 0.040$  & $134.6 \pm 2.5$ \\
\midrule
\multicolumn{5}{@{}l}{\emph{End of the streaming grid}} \\
 & path & context & energy, J & \\
\cmidrule(r){2-4}
TiRex             & incremental & 1\,048\,512 & $0.520 \pm 0.010$ & \\
TiRex-2           & incremental & 1\,048\,512 & $0.494 \pm 0.009$ & \\
Toto-2.0-22m      & incremental & 1\,048\,512 & $0.812 \pm 0.015$ & \\
Toto-1.0\tnote{\dag}         & incremental & 1\,048\,512 & $2.178 \pm 0.041$ & \\
TimesFM-2.5       & incremental &    498\,624 & $6.119 \pm 0.114$ & \\
Toto-2.0-22m      & recompute   & 1\,048\,512 & $991 \pm 19$ & \\
Chronos-2-small   & recompute   &    524\,160 & $3\,405 \pm 64$ & \\
\midrule
\multicolumn{5}{@{}l}{\emph{Power of the incremental refresh along the grid, W}} \\
 & 8\,256 & 123\,840 & 524\,160 & 1\,048\,512 \\
\cmidrule(r){2-5}
TiRex                      & $4.90 \pm 0.08$ & $4.98 \pm 0.08$ & $4.96 \pm 0.08$ & $5.03 \pm 0.07$ \\
TiRex-2                    & $4.78 \pm 0.09$ & $4.69 \pm 0.09$ & $4.71 \pm 0.10$ & $4.61 \pm 0.11$ \\
Toto-2.0-22m               & $4.41 \pm 0.11$ & $4.61 \pm 0.09$ & $4.59 \pm 0.09$ & $4.89 \pm 0.10$ \\
Toto-1.0\tnote{\dag}                   & $4.80 \pm 0.08$ & $5.03 \pm 0.09$ & $5.01 \pm 0.10$ & $4.99 \pm 0.08$ \\
TimesFM-2.5                & $4.88 \pm 0.09$ & $5.02 \pm 0.07$ & --              & --              \\
Toto-2.0-22m, recompute    & $6.03 \pm 0.20$ & $6.40 \pm 0.16$ & $6.69 \pm 0.10$ & $6.83 \pm 0.10$ \\
Chronos-2-small, recompute & $5.54 \pm 0.10$ & $5.80 \pm 0.13$ & $6.13 \pm 0.14$ & --              \\
\bottomrule
\end{tabular}
\end{table}

$^\dagger$Toto-1.0 is the only configuration with a sampling head. Every Toto-1.0 row in
this table times one sampled path, where the other configurations time one
deterministic forward pass; a deployment drawing the released default of $256$ paths
spends $256$ times the energy given here.

The single power per row holds only if the draw does not depend on the context. The third block
checks that: its step consumes one patch at every grid point, so the row has to be flat, and it is,
to $4.8\,\%$ over two decades. Only Toto-2.0-22m rises, by $10.9\,\%$ over the grid.

\newpage

\section{Additional Results}
\label{app:additionalresults}

\subsection{Streaming}
\label{app:streaming}

\noindent Throughout this section, columns are measurement cells and rows are the quantity and the
update path, named at the right edge; cells with nothing measured are dropped. Recompute and
incremental never share a panel, since on a linear axis they differ by three orders of magnitude.
Both axes are linear and every panel carries its own vertical range, so panel heights are not
comparable. A curve ends at the last context length measured, and a dot marks an end forced by the
memory or per-call time budget.

\begin{figure}[H]
\centering
\includegraphics{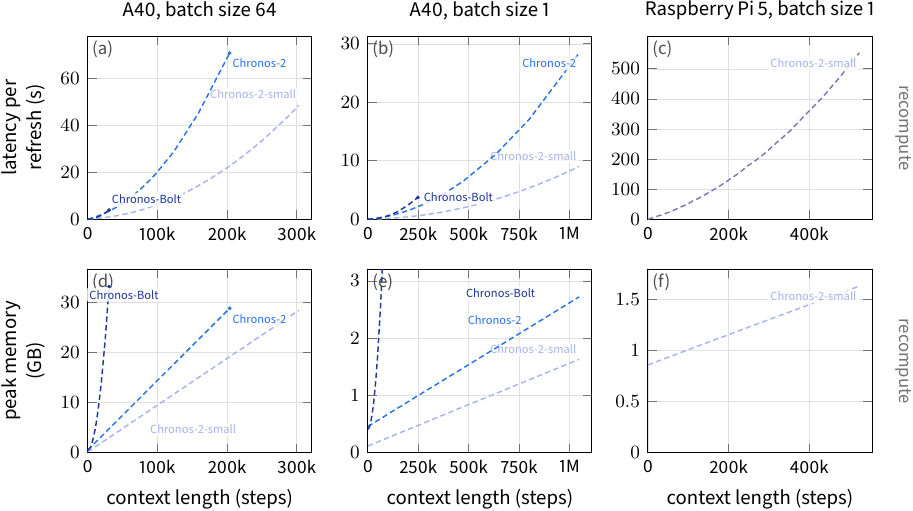}
\caption{The Chronos family, no incremental path. All three are non-causal encoders, so every point is a full re-encode and the comparison is one of constant and of growth rate, not of mechanism. Chronos-2 and Chronos-Bolt were measured on the A40 only. Chronos-Bolt is the outlier: its peak memory grows quadratically with the context, 32.9\,GB at batch size~1 and context 248\,448 where Chronos-2 needs 1.0\,GB. Its curve leaves panel~(e) at the top rather than compressing the other two.}
\label{fig:apx-chronos}
\end{figure}

\begin{figure}[H]
\centering
\includegraphics{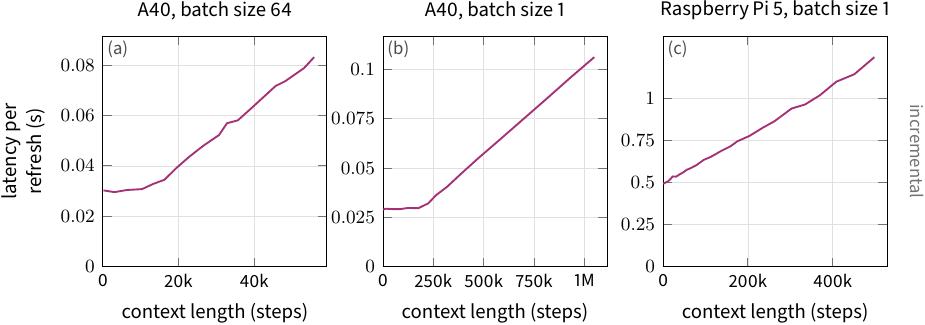}
\caption{TimesFM-2.5 on the cached path, the longest trained context in the set. Latency still grows with the context, which reflects attending to a cache rather than rebuilding it.}
\label{fig:apx-timesfm}
\end{figure}

\begin{figure}[H]
\centering
\includegraphics{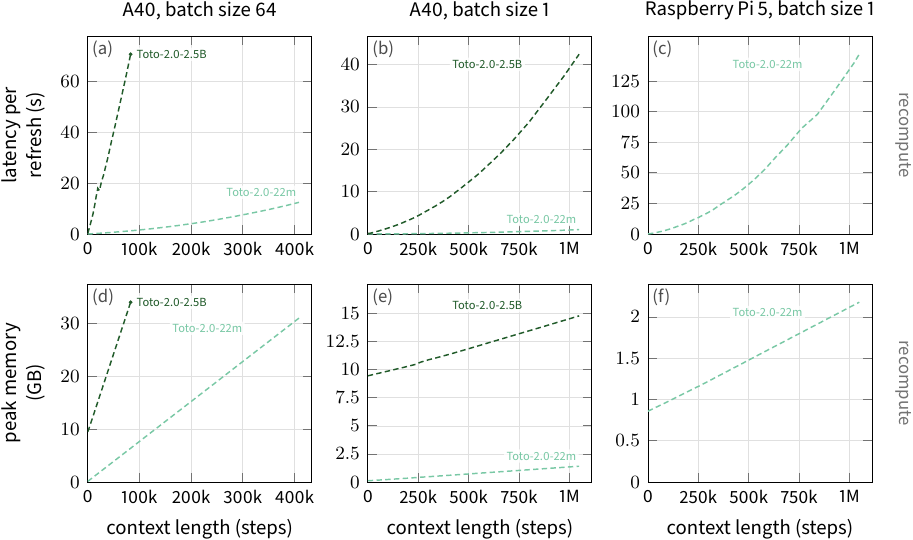}
\caption{Toto without a cache, at 22 and 2500 million parameters, each re-encoding the whole context at every step. Toto-2.0-2.5B is absent from the Raspberry\,Pi\,5 column: 9.36\,GB in fp32 does not fit the 7.9\,GiB available on that board.}
\label{fig:apx-toto-recompute}
\end{figure}

\begin{figure}[H]
\centering
\includegraphics{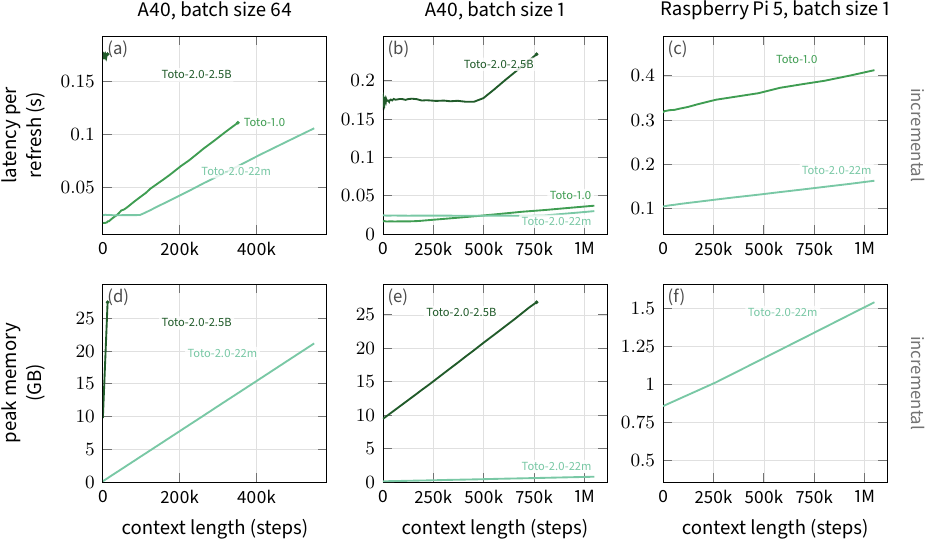}
\caption{Toto with the KV cache, now including Toto-1.0 (151\,M). Against the previous figure the shape changes, the ordering by size does not.}
\label{fig:apx-toto-cached}
\end{figure}

\begin{figure}[H]
\centering
\includegraphics{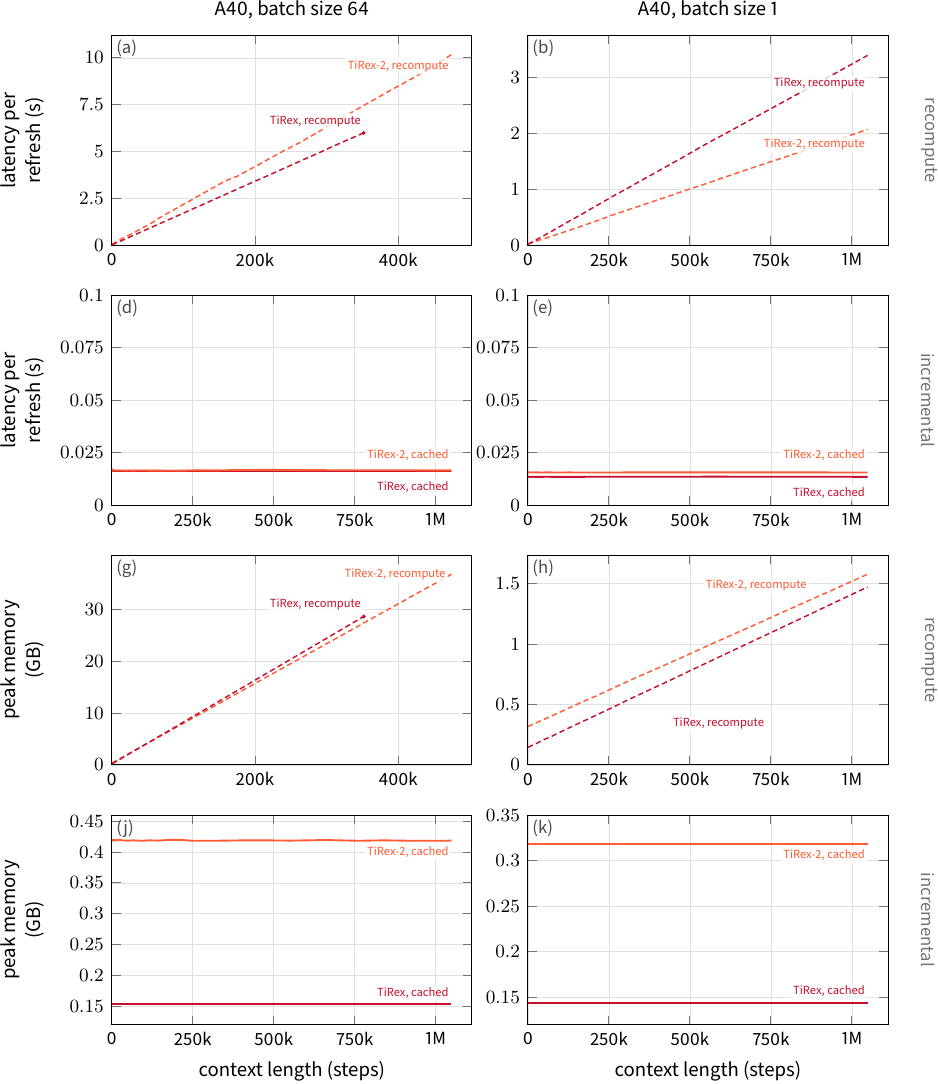}
\caption{TiRex against TiRex-2, dashed without and solid with the retained state. The solid rows are the only ones in the set that stay flat across the whole grid. Both memory rows carry both generations: on the incremental path TiRex holds 0.147\,GB at batch size~1 and 0.317\,GB at batch size~64, bit-identical across all 115 grid points, against 0.318 and 0.418\,GB for TiRex-2. The Raspberry\,Pi\,5 column is shown in Fig.~\ref{fig:apx-edge-regimes} instead. On the recompute path the two cross: TiRex is 0.80 times TiRex-2 at batch size~64 but 1.64 times at batch size~1 and context 1\,048\,512. Both recompute curves are flat below the trained context of 2048 and grow linearly above it.}
\label{fig:apx-tirex}
\end{figure}

\begin{figure}[H]
\centering
\includegraphics{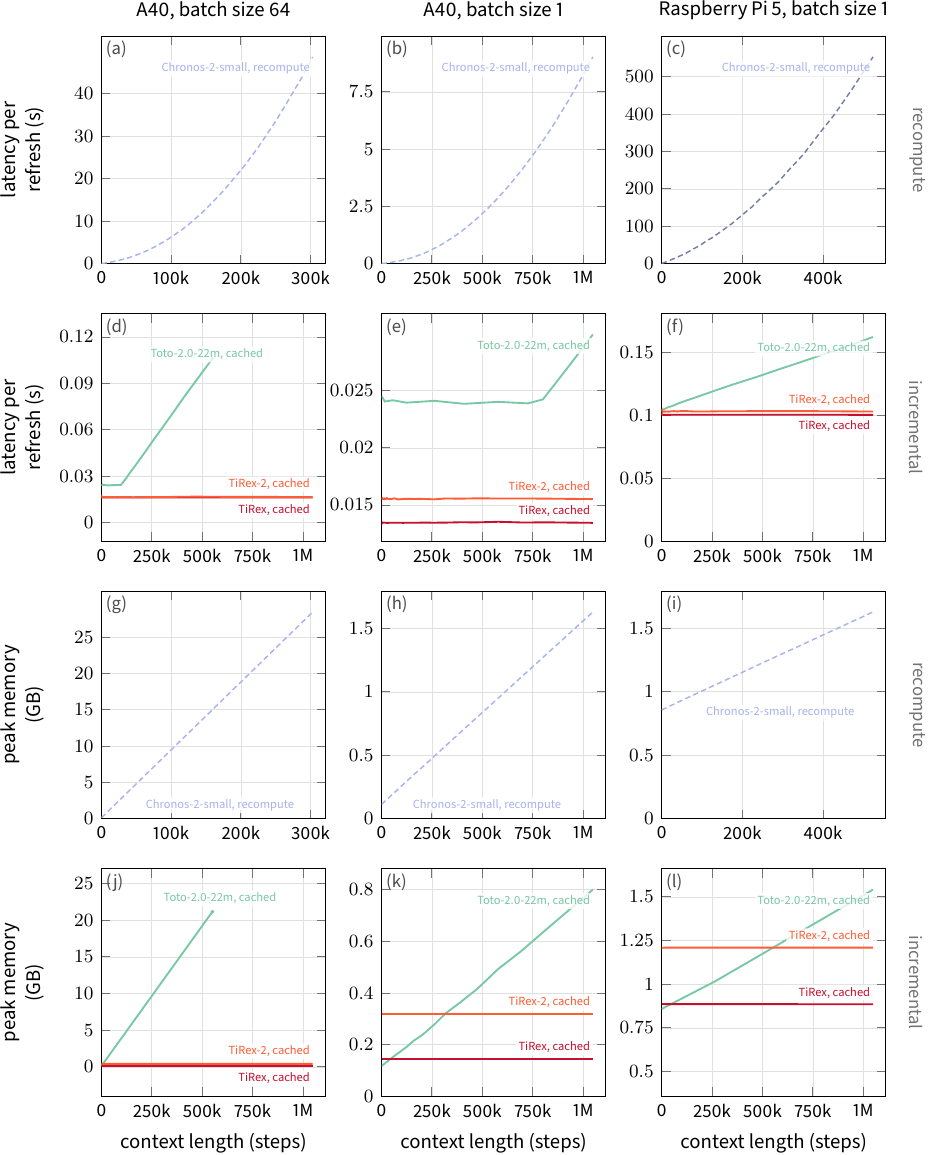}
\caption{Four architectures of comparable size at the lowest-latency path each admits: Chronos-2-small 28\,M, Toto-2.0-22m 22\,M, TiRex 35\,M and TiRex-2 82.5\,M active parameter.}
\label{fig:apx-matched-size}
\end{figure}

\begin{figure}[H]
\centering
\includegraphics{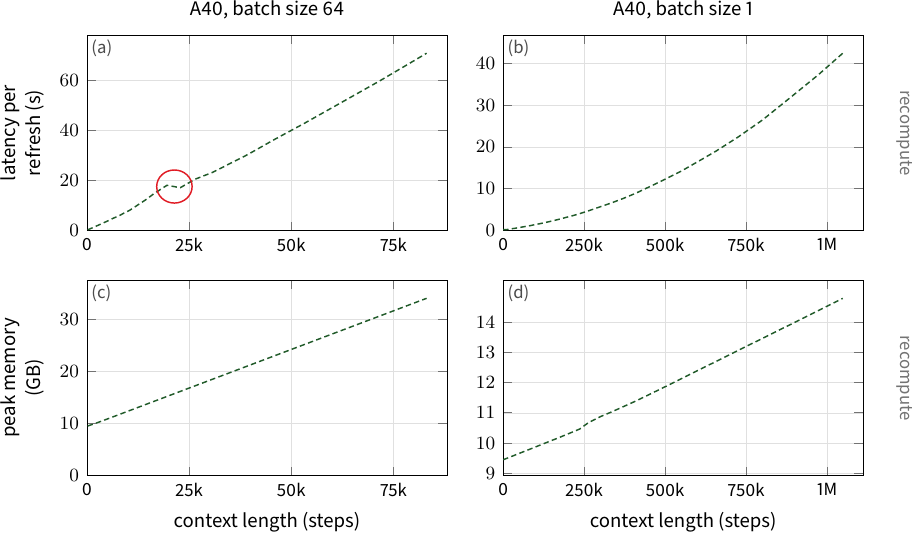}
\caption{Toto-2.0-2.5B without a cache, the highest latency and peak memory in the set. The ringed kink in panel~(a) is a kernel-selection artefact of the batch-64 fp32 call, not a property of the architecture; it is analysed in App.~\ref{app:sgemm-band}.}
\label{fig:apx-largest}
\end{figure}

\begin{figure}[H]
\centering
\includegraphics{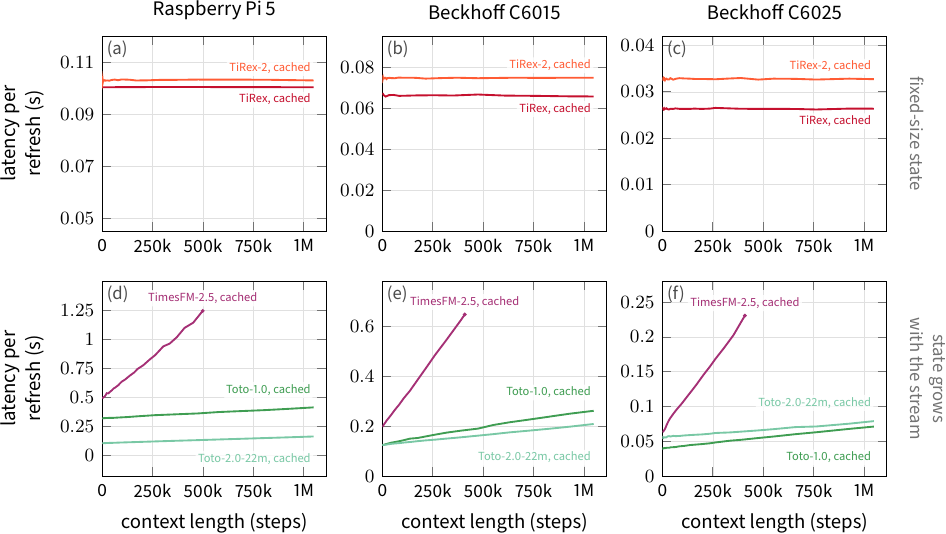}
\caption{Bounded against growing state on three edge machines, incremental path only. The upper row holds a fixed-size state, the lower a cache that grows with the stream. The split appears on all three boxes and does not follow their absolute speed.}
\label{fig:apx-edge-regimes}
\end{figure}

\begin{figure}[H]
\centering
\includegraphics{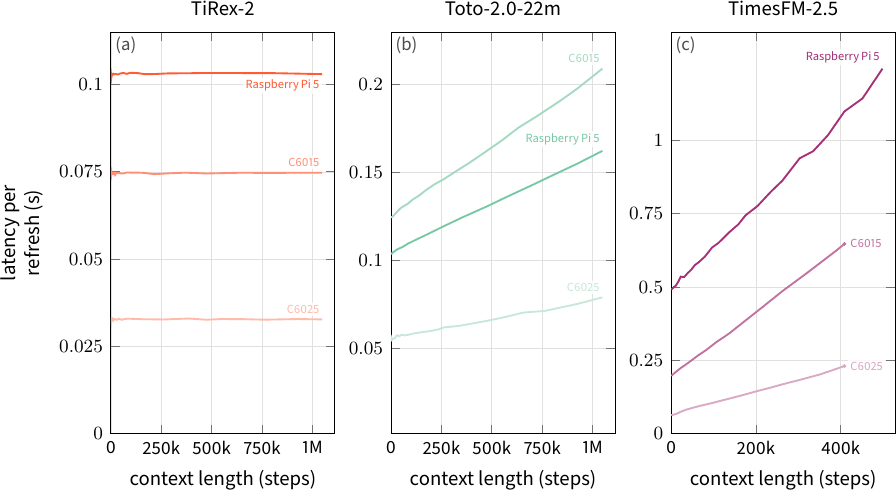}
\caption{One incremental path per column, one curve per machine, darkest Raspberry\,Pi\,5 to lightest C6025.}
\label{fig:apx-edge-portability}
\end{figure}

\begin{figure}[H]
\centering
\includegraphics{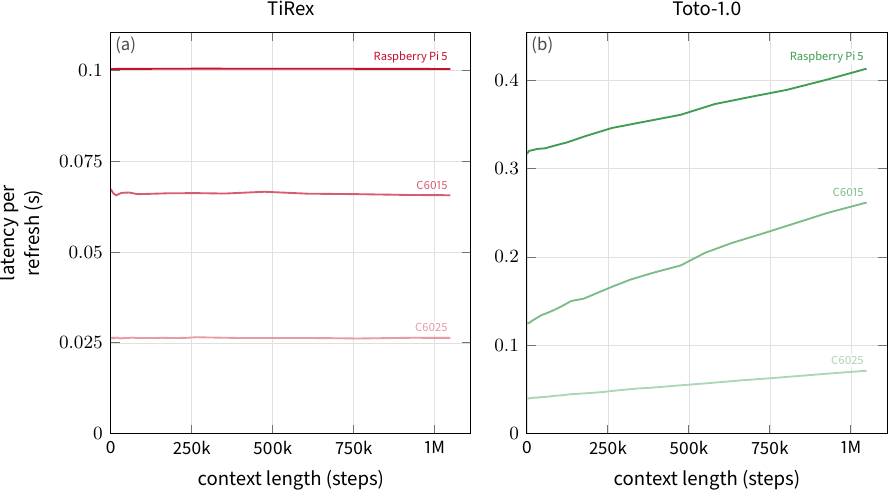}
\caption{The previous generation of each family that has one in the selected model group; same three machines and same shading.}
\label{fig:apx-edge-portability-prev}
\end{figure}
\newpage

\subsection{Accuracy}
\label{app:accuracy}

Table~\ref{tab:acc-agg} gives the aggregate behind Figure~\ref{fig:context-accuracy},
Tables~\ref{tab:acc-mixed} and~\ref{tab:acc-epf} the ten tasks individually. Skill is the
fev-bench skill score \citep{shchur_fev-bench_2025} of model $f$ over $R$ tasks against the
Seasonal Naive baseline $\beta$,
\begin{equation}
  \label{eq:skill}
  S_f = 100\left(1 - \left(\prod_{r=1}^{R}\operatorname{clip}\!\left(
  \frac{\mathrm{SQL}_{rf}}{\mathrm{SQL}_{r\beta}};\,10^{-2},\,10^{2}\right)\right)^{1/R}\right).
\end{equation}
One task is a single ratio, so a cell is $100\,(1-\mathrm{SQL}_f/\mathrm{SQL}_\beta)$ and an
aggregate averages ratios rather than scores. Runs are deterministic. Accuracy runs on an H100 NVL (94\,GB); \textsc{oom} is that budget.

\begin{table}[h!]
  \centering
  \caption{Skill score against context length, aggregated over the ten tasks whose series exceed 50\,000 steps. The column to read is \emph{stable to}: the longest context length whose score is at most two skill points below the score at the model's own trained context, and therefore the longest history the model may be handed without paying for it. \textbf{\texttt{full}} means the untruncated series qualifies. Individual shorter context lengths may dip below the threshold; only the endpoint is reported. $n/10$ marks an aggregate that would omit tasks. The best of the nine is set in bold in every context column. The \emph{shape} columns read the same window for every architecture, 2048 steps to the untruncated series: \emph{spread} is the largest minus the smallest score in it and \emph{drop} the largest fall from a running maximum. A flat row has both near zero; the best of the nine is set in bold in each.}
  \label{tab:acc-agg}
  \setlength{\tabcolsep}{2.0pt}
  \scriptsize
  \begin{tabular}{llrrrrrrrrrrrrrr}
    \toprule
    & trained & \multicolumn{11}{c}{context length given to the model (steps)} & stable & \multicolumn{2}{c}{shape, 2k--full} \\
    \cmidrule(lr){3-13}\cmidrule(lr){15-16}
    Architecture & context & 512 & 2k & 4k & 8k & 16k & 24k & 32k & 48k & 64k & 96k & full & to & spread & drop \\
    \midrule
    \multicolumn{16}{l}{\emph{Causal, recurrent: fixed-size state}} \\
    TiRex-2 & 8192 & \textbf{43.7} & 45.6 & 44.2 & 45.5 & \textbf{45.3} & \textbf{45.4} & \textbf{45.1} & \textbf{45.3} & 4/10 & 3/10 & \textbf{45.3} & \textbf{full} & \textbf{1.3} & \textbf{1.3} \\
    TiRex & 2048 & 34.9 & 37.0 & $-$33.4 & $-$60.8 & $-$70.6 & $-$80.8 & $-$94.7 & $-$75.8 & 4/10 & 3/10 & $-$75.3 & 2k & 131.8 & 131.8 \\
    \addlinespace[2pt]
    \multicolumn{16}{l}{\emph{Causal, attention: state grows with context}} \\
    Toto-1.0 & 4096 & 31.9 & 34.5 & 35.0 & 33.0 & 30.3 & 9/10 & 9/10 & 9/10 & 2/10 & 2/10 & 8/10 & 8k & 4.8 & 4.8 \\
    Toto-2.0-22m & 4096 & 32.9 & 36.3 & 36.9 & 32.5 & 34.1 & 33.5 & 32.0 & 29.2 & 4/10 & 3/10 & 27.2 & 4k & 9.7 & 9.7 \\
    Toto-2.0-2.5B & 4096 & 35.7 & 38.0 & 39.2 & 37.8 & 38.0 & 37.9 & 37.7 & 37.8 & 4/10 & 3/10 & 35.3 & 48k & 3.9 & 3.9 \\
    TimesFM-2.5 & 16256 & 31.5 & 31.7 & 33.8 & 34.0 & 34.8 & 32.8 & 32.3 & 25.5 & 4/10 & 3/10 & 8/10 & 24k & 9.3 & 9.3 \\
    \addlinespace[2pt]
    \multicolumn{16}{l}{\emph{Non-causal: no incremental state}} \\
    Chronos-2 & 8192 & 43.6 & \textbf{47.6} & \textbf{47.8} & \textbf{47.2} & 39.2 & 18.8 & $-$23.7 & $-$60.5 & 4/10 & 3/10 & $-$79.0 & 8k & 126.9 & 126.9 \\
    Chronos-2-small & 8192 & 42.8 & 45.3 & 45.7 & 45.2 & 15.7 & $-$24.9 & $-$60.5 & $-$99.2 & 4/10 & 3/10 & $-$108.5 & 8k & 154.3 & 154.3 \\
    Chronos-Bolt & 2048 & 33.0 & 34.8 & 33.3 & 30.6 & 16.6 & $-$57.2 & 9/10 & 9/10 & 3/10 & -- & 5/10 & 4k & 92.0 & 92.0 \\
    \bottomrule
  \end{tabular}
\end{table}

\begin{table}[p]
  \centering
  \caption{The five tasks that are not electricity prices; \emph{stable to} as in Table~\ref{tab:acc-agg}, which also gives the trained context of each architecture. Markers: \textsc{oom} out of memory, \textsc{sdpa} the 65\,535 batch limit of efficient attention, \textsc{err} another failure, -- not measured. In the \texttt{full} column the best of the nine is set in bold, as are the best \emph{spread} and \emph{drop} of each task.}
  \label{tab:acc-mixed}
  \setlength{\tabcolsep}{2.0pt}
  \scriptsize
  \begin{tabular}{lrrrrrrrrrrrrrr}
    \toprule
    & \multicolumn{11}{c}{context length given to the model (steps)} & stable & \multicolumn{2}{c}{shape, 2k--full} \\
    \cmidrule(lr){2-12}\cmidrule(lr){14-15}
    Architecture & 512 & 2k & 4k & 8k & 16k & 24k & 32k & 48k & 64k & 96k & full & to & spread & drop \\
    \midrule
    \multicolumn{15}{l}{\texttt{solar\_with\_weather\_15T} --- full context 198\,504 steps} \\
    TiRex-2 & 32.8 & 32.6 & 33.4 & 35.7 & 34.8 & 35.9 & 35.1 & 35.7 & 34.8 & 35.2 & \textbf{35.5} & \textbf{full} & \textbf{3.3} & \textbf{1.2} \\
    TiRex & 29.6 & 29.2 & $-$73.7 & $-$18.3 & $-$104.1 & $-$28.6 & $-$34.1 & $-$20.1 & $-$30.4 & $-$30.5 & $-$29.7 & 2k & 133.3 & 133.3 \\
    Toto-1.0 & 25.2 & 35.9 & 34.5 & 33.1 & 26.9 & 26.6 & 25.3 & 24.5 & 25.0 & 21.9 & 7.2 & 8k & 28.7 & 28.7 \\
    Toto-2.0-22m & 27.8 & 33.6 & 31.7 & 33.8 & 31.9 & 28.9 & 28.3 & 32.9 & 29.9 & 31.0 & 32.5 & \textbf{full} & 5.5 & 5.5 \\
    Toto-2.0-2.5B & 30.6 & 33.2 & 37.9 & 35.0 & 34.7 & 34.6 & 33.6 & 34.7 & 33.5 & 33.2 & 33.2 & 4k & 4.7 & 4.7 \\
    TimesFM-2.5 & 30.3 & 18.6 & 24.9 & 22.4 & 24.1 & 30.0 & 30.3 & 30.6 & 32.0 & 25.6 & \textsc{oom} & 96k & 13.5 & 6.4 \\
    Chronos-2 & 38.0 & 44.6 & 44.5 & 43.3 & 31.3 & 11.4 & $-$40.0 & $-$52.9 & $-$52.7 & $-$50.7 & $-$53.0 & 8k & 98.0 & 98.0 \\
    Chronos-2-small & 36.1 & 39.9 & 41.5 & 40.7 & 10.6 & $-$27.3 & $-$45.4 & $-$57.6 & $-$55.5 & $-$54.0 & $-$70.9 & 8k & 112.4 & 112.4 \\
    Chronos-Bolt & 25.3 & 33.3 & 32.9 & 29.3 & 22.0 & $-$3.8 & $-$30.2 & $-$42.4 & $-$44.8 & \textsc{oom} & \textsc{oom} & 4k & 78.1 & 78.1 \\
    \addlinespace[3pt]
    \multicolumn{15}{l}{\texttt{ercot\_1H} --- full context 154\,704 steps} \\
    TiRex-2 & 16.6 & 14.9 & 12.4 & 13.7 & 13.1 & 14.1 & 13.7 & 13.1 & 12.7 & 12.1 & \textbf{12.5} & \textbf{full} & \textbf{3.9} & \textbf{3.9} \\
    TiRex & 19.2 & 19.4 & $-$159.1 & $-$180.0 & $-$200.0 & $-$225.1 & $-$236.1 & $-$319.7 & $-$379.9 & $-$491.1 & $-$575.6 & 2k & 595.0 & 595.0 \\
    Toto-1.0 & 16.4 & 13.4 & 17.0 & 11.3 & 8.4 & 6.5 & 5.9 & 2.9 & \textsc{sdpa} & $-$1.9 & \textsc{sdpa} & 6k & 19.6 & 19.6 \\
    Toto-2.0-22m & 10.2 & 16.0 & 16.8 & $-$30.1 & $-$11.7 & $-$13.2 & $-$15.7 & $-$13.1 & $-$10.6 & $-$14.2 & $-$23.4 & 4k & 47.8 & 47.8 \\
    Toto-2.0-2.5B & 19.0 & 19.5 & 18.8 & 11.5 & 9.5 & 11.6 & 11.5 & 11.8 & 8.6 & 1.6 & $-$27.0 & 4k & 46.4 & 46.4 \\
    TimesFM-2.5 & $-$4.0 & 7.3 & 9.1 & 11.0 & 12.8 & 5.9 & 0.3 & $-$10.4 & 11.0 & $-$19.3 & \textsc{oom} & 64k & 32.3 & 32.3 \\
    Chronos-2 & 21.0 & 21.1 & 22.2 & 22.1 & 13.4 & $-$20.2 & $-$117.0 & $-$173.7 & $-$242.9 & $-$355.4 & $-$457.9 & 12k & 480.6 & 480.6 \\
    Chronos-2-small & 17.5 & 16.9 & 18.7 & 19.2 & $-$3.0 & $-$99.3 & $-$176.9 & $-$277.1 & $-$307.8 & $-$372.9 & $-$454.6 & 8k & 474.1 & 474.1 \\
    Chronos-Bolt & 13.3 & 15.6 & 14.9 & 11.3 & 2.0 & $-$47.4 & $-$113.6 & $-$191.1 & $-$222.0 & \textsc{oom} & \textsc{oom} & 4k & 239.8 & 239.8 \\
    \addlinespace[3pt]
    \multicolumn{15}{l}{\texttt{LOOP\_SEATTLE\_5T} --- full context 104\,832 steps} \\
    TiRex-2 & 24.3 & 26.6 & 22.6 & 23.8 & 24.0 & 23.6 & 23.7 & 23.9 & 23.9 & 24.1 & \textbf{24.3} & \textbf{full} & \textbf{4.5} & \textbf{4.5} \\
    TiRex & 22.7 & 27.0 & $-$111.7 & $-$106.0 & $-$96.7 & $-$82.8 & $-$76.0 & $-$78.2 & $-$82.6 & $-$76.8 & $-$75.1 & 2k & 157.0 & 157.0 \\
    Toto-1.0 & 19.0 & 26.4 & 25.4 & 25.1 & 20.6 & \textsc{sdpa} & \textsc{sdpa} & \textsc{sdpa} & \textsc{sdpa} & \textsc{sdpa} & \textsc{sdpa} & 8k & 6.1 & 6.1 \\
    Toto-2.0-22m & 11.5 & 27.4 & 27.8 & 24.5 & 27.4 & 26.1 & 25.8 & 24.1 & 22.5 & 20.7 & 20.6 & 32k & 9.6 & 9.6 \\
    Toto-2.0-2.5B & 13.5 & 25.3 & 25.1 & 25.9 & 22.7 & 22.1 & 22.1 & 21.6 & 21.6 & 21.7 & 21.9 & 12k & 4.6 & 4.6 \\
    TimesFM-2.5 & 22.4 & 21.0 & 20.8 & 20.8 & 20.8 & 21.3 & 21.3 & 22.7 & 22.5 & 17.0 & 18.9 & \textbf{full} & 5.7 & 5.7 \\
    Chronos-2 & 14.3 & 29.7 & 27.8 & 29.1 & 21.5 & 11.0 & $-$1.6 & $-$14.9 & $-$20.5 & $-$27.3 & $-$27.3 & 8k & 57.1 & 57.1 \\
    Chronos-2-small & 15.4 & 28.7 & 25.5 & 26.9 & 5.1 & $-$4.6 & $-$18.8 & $-$43.6 & $-$45.0 & $-$56.0 & $-$57.9 & 12k & 86.6 & 86.6 \\
    Chronos-Bolt & 11.6 & 15.3 & 5.7 & $-$4.5 & $-$12.2 & $-$36.4 & $-$73.2 & $-$132.5 & $-$159.5 & \textsc{oom} & \textsc{oom} & 2k & 174.8 & 174.8 \\
    \addlinespace[3pt]
    \multicolumn{15}{l}{\texttt{ETT\_15T} --- full context 69\,584 steps} \\
    TiRex-2 & 18.9 & 26.7 & 25.2 & 26.6 & 26.4 & 26.4 & 26.2 & 25.9 & 25.9 & -- & \textbf{25.7} & \textbf{full} & \textbf{1.9} & \textbf{1.5} \\
    TiRex & 18.2 & 25.5 & $-$42.2 & $-$67.5 & $-$88.5 & $-$166.1 & $-$89.3 & $-$204.2 & $-$91.1 & -- & $-$128.3 & 2k & 229.7 & 229.7 \\
    Toto-1.0 & 14.2 & 21.4 & 22.3 & 21.4 & 17.2 & 14.1 & 13.3 & 8.9 & 5.3 & -- & 5.8 & 8k & 17.5 & 17.5 \\
    Toto-2.0-22m & 19.9 & 26.1 & 27.7 & 25.1 & 25.0 & 22.2 & 20.1 & 16.2 & 14.1 & -- & 13.2 & 12k & 14.5 & 14.5 \\
    Toto-2.0-2.5B & 19.6 & 27.1 & 30.0 & 28.9 & 27.6 & 26.7 & 26.7 & 25.6 & 25.3 & -- & 25.2 & 12k & 4.8 & 4.8 \\
    TimesFM-2.5 & 20.1 & 22.3 & 24.4 & 24.3 & 24.2 & 23.1 & 24.4 & 20.9 & 22.3 & -- & 21.3 & 64k & 3.6 & 3.6 \\
    Chronos-2 & 21.9 & 27.0 & 28.3 & 28.4 & 15.7 & $-$10.7 & $-$41.3 & $-$74.6 & $-$101.7 & -- & $-$106.2 & 8k & 134.6 & 134.6 \\
    Chronos-2-small & 22.2 & 25.0 & 25.0 & 24.8 & $-$4.9 & $-$44.4 & $-$79.8 & $-$139.1 & $-$139.2 & -- & $-$134.5 & 8k & 164.3 & 164.3 \\
    Chronos-Bolt & 20.6 & 26.3 & 24.8 & 21.9 & 3.3 & $-$22.4 & $-$50.6 & $-$531.4 & \textsc{oom} & -- & \textsc{oom} & 4k & 557.7 & 557.7 \\
    \addlinespace[3pt]
    \multicolumn{15}{l}{\texttt{jena\_weather\_10T} --- full context 52\,560 steps} \\
    TiRex-2 & 40.4 & 43.9 & 44.5 & 45.3 & 43.0 & 45.2 & 45.1 & 45.0 & -- & -- & 44.7 & \textbf{full} & \textbf{2.3} & \textbf{2.3} \\
    TiRex & 35.2 & 42.1 & $-$17.6 & $-$58.4 & $-$72.2 & $-$60.7 & $-$104.7 & $-$28.2 & -- & -- & $-$29.4 & 2k & 146.9 & 146.9 \\
    Toto-1.0 & 40.6 & 44.2 & 45.4 & 42.8 & 42.7 & 42.1 & 40.9 & 39.2 & -- & -- & 14.2 & 12k & 31.2 & 31.2 \\
    Toto-2.0-22m & 40.1 & 44.7 & 47.7 & 37.4 & 40.3 & 36.7 & 31.1 & 28.8 & -- & -- & 27.2 & 4k & 20.4 & 20.4 \\
    Toto-2.0-2.5B & 39.5 & 46.2 & 48.7 & 50.1 & 50.9 & 49.5 & 48.2 & 48.5 & -- & -- & \textbf{48.6} & \textbf{full} & 4.7 & 2.7 \\
    TimesFM-2.5 & 36.2 & 42.3 & 44.1 & 46.1 & 47.0 & 43.5 & 44.2 & 44.1 & -- & -- & 44.4 & 16k & 4.7 & 3.6 \\
    Chronos-2 & 42.6 & 48.5 & 48.1 & 47.4 & 42.3 & 24.2 & 2.2 & $-$20.3 & -- & -- & $-$22.9 & 8k & 71.4 & 71.4 \\
    Chronos-2-small & 39.0 & 46.4 & 46.7 & 46.5 & 27.2 & 11.4 & $-$43.9 & $-$61.3 & -- & -- & $-$56.9 & 8k & 108.2 & 108.2 \\
    Chronos-Bolt & 33.3 & 41.0 & 42.0 & 38.8 & 12.0 & $-$3114.2 & \textsc{oom} & \textsc{oom} & -- & -- & \textsc{oom} & 6k & 3156.1 & 3156.1 \\
    \bottomrule
  \end{tabular}
\end{table}

\begin{table}[p]
  \centering
  \caption{The five electricity-price tasks, layout and markers as in Table~\ref{tab:acc-mixed}. They share length, horizon and variates and each rests on a single series.}
  \label{tab:acc-epf}
  \setlength{\tabcolsep}{2.0pt}
  \scriptsize
  \begin{tabular}{lrrrrrrrrrrrr}
    \toprule
    & \multicolumn{9}{c}{context length given to the model (steps)} & stable & \multicolumn{2}{c}{shape, 2k--full} \\
    \cmidrule(lr){2-10}\cmidrule(lr){12-13}
    Architecture & 512 & 2k & 4k & 8k & 16k & 24k & 32k & 48k & full & to & spread & drop \\
    \midrule
    \multicolumn{13}{l}{\texttt{epf\_be} --- full context 52\,392 steps} \\
    TiRex-2 & 53.2 & 53.8 & 52.1 & 52.2 & 52.6 & 52.1 & 52.7 & 52.9 & 53.1 & \textbf{full} & \textbf{1.7} & \textbf{1.7} \\
    TiRex & 49.6 & 54.2 & 23.6 & $-$57.6 & $-$26.6 & $-$28.0 & $-$22.4 & $-$4.6 & 16.8 & 2k & 111.8 & 111.8 \\
    Toto-1.0 & 51.2 & 51.7 & 51.2 & 48.2 & 43.3 & 43.5 & 41.7 & 39.6 & 39.0 & 6k & 12.7 & 12.7 \\
    Toto-2.0-22m & 52.2 & 51.8 & 52.9 & 51.2 & 51.6 & 51.9 & 51.3 & 49.5 & 46.7 & 32k & 6.2 & 6.2 \\
    Toto-2.0-2.5B & 55.5 & 54.5 & 55.0 & 54.6 & 55.4 & 55.5 & 54.6 & 54.0 & \textbf{53.8} & \textbf{full} & 1.8 & 1.8 \\
    TimesFM-2.5 & 47.7 & 54.1 & 56.1 & 55.9 & 57.0 & 57.1 & 55.3 & 52.9 & 51.4 & 32k & 5.7 & 5.7 \\
    Chronos-2 & 50.2 & 54.4 & 56.3 & 56.2 & 51.1 & 42.4 & 21.3 & $-$18.8 & $-$21.9 & 12k & 78.2 & 78.2 \\
    Chronos-2-small & 54.4 & 54.1 & 55.1 & 54.9 & 31.9 & 3.0 & $-$9.8 & $-$30.7 & $-$30.2 & 8k & 85.8 & 85.8 \\
    Chronos-Bolt & 50.8 & 50.2 & 51.8 & 49.5 & 32.4 & 17.7 & $-$2.5 & $-$56.7 & $-$96.6 & 8k & 148.4 & 148.4 \\
    \addlinespace[3pt]
    \multicolumn{13}{l}{\texttt{epf\_de} --- full context 52\,392 steps} \\
    TiRex-2 & 60.1 & 64.9 & 65.8 & 65.0 & 65.6 & 65.5 & 65.3 & 65.3 & \textbf{65.5} & \textbf{full} & \textbf{1.4} & \textbf{0.7} \\
    TiRex & 28.6 & 25.6 & $-$17.1 & $-$17.9 & $-$26.9 & $-$31.3 & $-$32.9 & $-$28.7 & $-$28.6 & 2k & 58.5 & 58.5 \\
    Toto-1.0 & 20.1 & 23.6 & 20.9 & 17.8 & 16.4 & 16.7 & 16.5 & 16.6 & 16.2 & 6k & 7.4 & 7.4 \\
    Toto-2.0-22m & 26.8 & 25.5 & 25.7 & 26.1 & 25.8 & 27.0 & 27.8 & 28.0 & 27.6 & \textbf{full} & 2.6 & 1.2 \\
    Toto-2.0-2.5B & 27.6 & 25.9 & 27.2 & 27.3 & 27.7 & 27.7 & 29.0 & 30.6 & 30.8 & \textbf{full} & 4.8 & 1.9 \\
    TimesFM-2.5 & 17.6 & 25.2 & 25.6 & 24.9 & 25.8 & 13.6 & 14.3 & $-$1.0 & 4.2 & 16k & 26.8 & 26.8 \\
    Chronos-2 & 59.4 & 65.9 & 65.5 & 64.6 & 54.5 & 22.1 & $-$19.1 & $-$24.5 & $-$25.8 & 8k & 91.7 & 91.7 \\
    Chronos-2-small & 61.7 & 65.8 & 66.2 & 65.9 & 18.4 & $-$14.9 & $-$27.4 & $-$41.1 & $-$41.1 & 8k & 107.3 & 107.3 \\
    Chronos-Bolt & 26.7 & 26.4 & 22.9 & 20.8 & 12.8 & $-$19.2 & $-$73.3 & $-$117.7 & $-$104.6 & 2k & 144.1 & 144.1 \\
    \addlinespace[3pt]
    \multicolumn{13}{l}{\texttt{epf\_fr} --- full context 52\,392 steps} \\
    TiRex-2 & 71.7 & 72.0 & 71.1 & 71.8 & 71.8 & 71.7 & 71.8 & 71.7 & \textbf{71.7} & \textbf{full} & 1.0 & 1.0 \\
    TiRex & 67.6 & 67.8 & 44.0 & $-$75.9 & $-$81.6 & $-$96.6 & $-$86.4 & 17.5 & 10.0 & 2k & 164.4 & 164.4 \\
    Toto-1.0 & 65.9 & 64.7 & 65.9 & 63.5 & 62.4 & 61.1 & 60.2 & 57.8 & 57.8 & 6k & 8.1 & 8.1 \\
    Toto-2.0-22m & 64.9 & 65.4 & 65.0 & 65.2 & 64.7 & 65.9 & 65.0 & 63.0 & 63.6 & \textbf{full} & 2.8 & 2.8 \\
    Toto-2.0-2.5B & 67.2 & 67.3 & 68.2 & 67.5 & 68.2 & 68.2 & 68.0 & 67.7 & 67.6 & \textbf{full} & \textbf{0.9} & \textbf{0.7} \\
    TimesFM-2.5 & 67.1 & 65.5 & 66.7 & 66.9 & 67.2 & 64.7 & 66.0 & 56.0 & 57.9 & 32k & 11.3 & 11.3 \\
    Chronos-2 & 71.2 & 71.2 & 71.1 & 71.0 & 67.2 & 57.4 & 21.5 & $-$42.1 & $-$42.0 & 8k & 113.7 & 113.7 \\
    Chronos-2-small & 71.2 & 68.0 & 67.4 & 67.2 & 50.3 & 2.7 & $-$33.3 & $-$74.4 & $-$71.6 & 8k & 142.7 & 142.7 \\
    Chronos-Bolt & 65.7 & 64.8 & 65.2 & 63.9 & 49.9 & 26.6 & $-$39.9 & $-$227.0 & $-$232.3 & 8k & 297.5 & 297.5 \\
    \addlinespace[3pt]
    \multicolumn{13}{l}{\texttt{epf\_np} --- full context 52\,392 steps} \\
    TiRex-2 & 59.0 & 60.0 & 55.1 & 60.2 & 60.2 & 58.5 & 57.3 & 58.3 & \textbf{58.3} & \textbf{full} & 6.1 & 6.1 \\
    TiRex & 36.9 & 36.8 & $-$16.2 & $-$58.2 & $-$92.8 & $-$299.7 & $-$457.2 & $-$518.8 & $-$493.6 & 2k & 555.6 & 555.6 \\
    Toto-1.0 & 24.4 & 29.1 & 32.1 & 29.1 & 28.2 & 25.9 & 24.4 & 23.9 & 24.0 & 12k & 8.2 & 8.2 \\
    Toto-2.0-22m & 34.2 & 35.2 & 35.8 & 35.5 & 34.2 & 34.7 & 32.5 & 17.8 & 13.7 & 24k & 22.8 & 22.8 \\
    Toto-2.0-2.5B & 38.9 & 38.5 & 37.8 & 35.9 & 37.0 & 37.4 & 38.1 & 38.0 & 37.9 & \textbf{full} & \textbf{2.5} & \textbf{2.5} \\
    TimesFM-2.5 & 31.9 & 21.5 & 22.7 & 22.3 & 22.9 & 23.5 & 19.2 & 10.7 & 11.4 & 24k & 12.8 & 12.8 \\
    Chronos-2 & 59.9 & 58.1 & 58.5 & 57.0 & 43.9 & $-$8.2 & $-$101.0 & $-$168.6 & $-$187.3 & 8k & 245.8 & 245.8 \\
    Chronos-2-small & 55.1 & 55.8 & 57.5 & 55.1 & 0.2 & $-$73.6 & $-$177.4 & $-$370.8 & $-$364.9 & 8k & 428.3 & 428.3 \\
    Chronos-Bolt & 38.7 & 36.6 & 30.2 & 24.3 & 7.5 & $-$41.9 & $-$135.5 & $-$231.3 & $-$245.6 & 2k & 282.2 & 282.2 \\
    \addlinespace[3pt]
    \multicolumn{13}{l}{\texttt{epf\_pjm} --- full context 52\,392 steps} \\
    TiRex-2 & 27.9 & 27.2 & 27.2 & 27.2 & 27.5 & 27.8 & 27.8 & 28.0 & \textbf{28.6} & \textbf{full} & \textbf{1.5} & \textbf{0.8} \\
    TiRex & 20.1 & 21.6 & $-$92.8 & $-$23.7 & 5.1 & 13.4 & $-$50.9 & $-$17.4 & $-$5.5 & 2k & 117.4 & 117.4 \\
    Toto-1.0 & 17.4 & 12.2 & 12.5 & 17.7 & 16.1 & 15.5 & 14.1 & 13.9 & 13.7 & \textbf{full} & 6.0 & 4.5 \\
    Toto-2.0-22m & 16.2 & 17.7 & 18.2 & 20.4 & 25.1 & 25.7 & 24.9 & 17.0 & 17.4 & \textbf{full} & 8.8 & 8.8 \\
    Toto-2.0-2.5B & 21.0 & 21.8 & 21.5 & 16.8 & 19.5 & 20.5 & 21.0 & 21.9 & 21.9 & \textbf{full} & 5.1 & 5.1 \\
    TimesFM-2.5 & 18.1 & 11.3 & 15.5 & 17.9 & 17.6 & 16.9 & 16.3 & $-$7.9 & $-$6.9 & 32k & 27.1 & 27.1 \\
    Chronos-2 & 25.5 & 28.2 & 27.9 & 25.9 & 25.3 & 23.8 & $-$33.5 & $-$98.9 & $-$95.8 & 16k & 127.8 & 127.8 \\
    Chronos-2-small & 20.9 & 24.3 & 24.8 & 24.3 & 1.5 & $-$44.6 & $-$77.6 & $-$97.8 & $-$92.7 & 8k & 122.6 & 122.6 \\
    Chronos-Bolt & 21.0 & 18.1 & 17.9 & 23.9 & 18.8 & $-$9.8 & $-$64.5 & $-$859.8 & $-$899.9 & 16k & 923.9 & 923.9 \\
    \bottomrule
  \end{tabular}
\end{table}

\newpage
\section{Kernel selection in the Toto-2.0-2.5B curve}
\label{app:sgemm-band}

\paragraph{Observation.}
In Experiment 3 (Sec.~\ref{sec:streaming}) and contrary to expectation, Toto-2.0-2.5B with a batch size of $64$ is faster at a
context of $\num{22848}$ steps than at $\num{20736}$, as can be seen in Fig.~\ref{fig:apx-largest}(a) in
App.~\ref{app:streaming}. The difference is $17.0$\,s against $19.5$\,s for a $10\%$ longer context.

\paragraph{Cause.}
Peak memory is linear in token count across the whole grid and
identical to the byte across runs, which means nothing is being truncated. Rerunning the two
grid points alone in a fresh process on an idle device, ascending and
descending, reproduces both timings within $1\%$, ruling out grid order,
fragmentation, thermal effects and competing load. Attention accounts for under
$6\%$ of the call at either point.

\begin{figure}[H]
\centering
\includegraphics{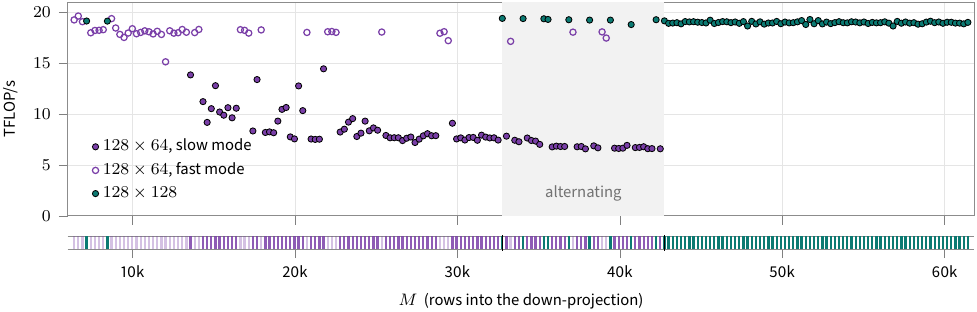}
\caption{The MLP down-projection measured on its own, fp32 with TF32 disabled,
A40, $M$ from 6400 to 61440 in steps of 256. \textbf{Top:} throughput. The
$128\times128$ tile is flat at $19.0$\,TFLOP/s wherever it is chosen; the
$128\times64$ tile matches it only up to $M\approx13568$ and then falls to a
median of $7.7$, with recoveries that do not persist. We call the two levels
fast and slow mode, split at $15$\,TFLOP/s; no measured point lies between
$14.5$ and $15.1$. \textbf{Bottom:} which tile cuBLAS selected, one bar per grid
point on the same axis. The two black marks bound the interval in which the
choice alternates from one point to the next. Reading the panels together, the
throughput of a call is set by the bar below it rather than by $M$.}
\label{fig:sgemm-scan}
\end{figure}

What changes is the kernel cuBLAS picks for one matrix multiplication, the MLP
down-projection, which sees $M=\mathrm{batch}\times(\text{tokens}+1)$ rows. Two
kernels are available: one computes the output in $128\times128$ tiles, the
other in $128\times64$. The wider tile holds $19.0$\,TFLOP/s at every $M$ at
which it is chosen, while the narrower one drops to a median of $7.7$ above
$M\approx13568$ (Fig.~\ref{fig:sgemm-scan}). Between the two grid points,
$M=41536$ and $M=45760$, cuBLAS switches from the narrow tile to the wide one in
all 48 layers, which is how the longer context came to be served faster. The
selection is specific to the cuBLAS shipped with PyTorch 2.9.1 (CUDA 12.8) under driver 590.48.01.

\begin{figure}[t]
\centering
\includegraphics{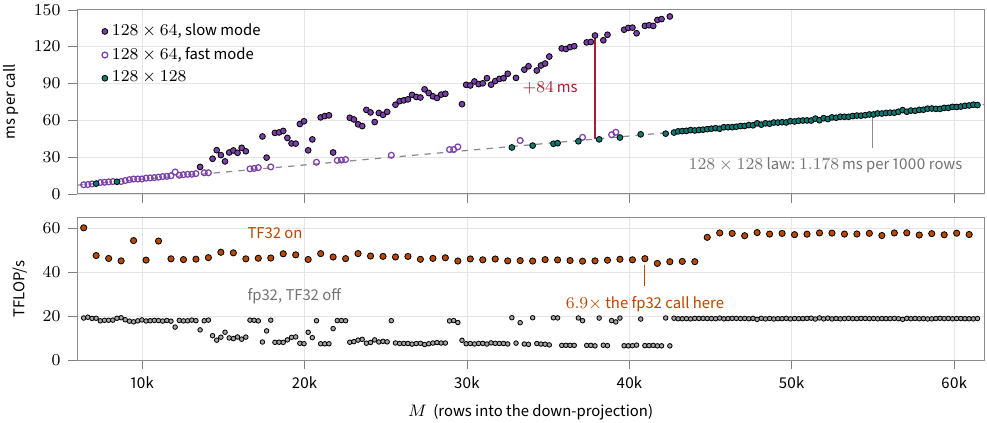}
\caption{\textbf{Top:}~the calls of Fig.~\ref{fig:sgemm-scan} as time rather than
throughput, against the law the $128\times128$ regime obeys. The distance above
it is the excess, marked at $M=37888$; forty-eight layers multiply it into the
seconds visible in the streaming curve. \textbf{Bottom:}~the same scan with
tensor-core kernels enabled, over every third grid point, against the fp32
measurements in grey. TF32 does not collapse with $M$, but it is not flat either:
it steps from about $46$ to about $57$\,TFLOP/s near $M=44800$, shortly above the
fp32 switch point.}
\label{fig:sgemm-cost}
\end{figure}

Figure~\ref{fig:sgemm-cost}, top, gives the same calls as time against the
$128\times128$ law, $1.178$\,ms per 1000 rows. The excess reaches $94$\,ms at
$M=42496$, $2.9\times$ the predicted time, and grows with $M$ as slow-mode
throughput falls from $10.3$ to $6.6$\,TFLOP/s; subtracting $48$ times the
per-layer difference recovers the linear law to $0.5\%$ at context 20736.

\begin{figure}[t]
\centering
\includegraphics{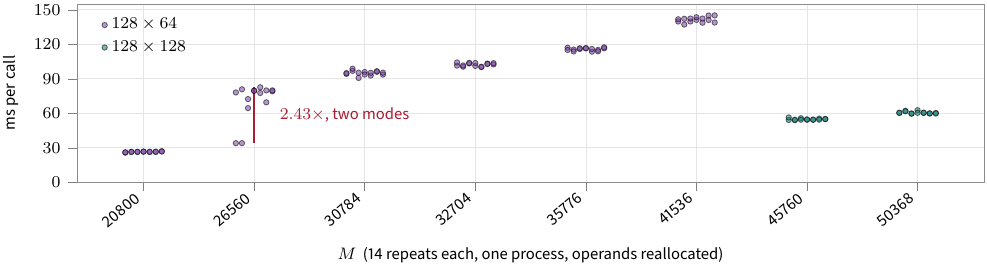}
\caption{Fourteen repeats at eight fixed $M$, jittered horizontally. Seven series
are tight; $M=26560$ splits cleanly in two. The mode is selected once per
allocation and then held, which is why a placement-dependent effect reproduces
across runs. The sixth and seventh series are the two grid points of
Fig.~\ref{fig:apx-largest}(a).}
\label{fig:sgemm-modes}
\end{figure}

The slow mode is not a property of the shape alone.
Figure~\ref{fig:sgemm-modes} repeats the multiplication 14 times at eight fixed
$M$: seven series are tight, $M=26560$ splits into $34$ and $79$\,ms, and
$M=20800$ runs fast where the scan had it slow. The mode is therefore fixed per
allocation, which is why the band reproduces across runs; we did not confirm the
mechanism with memory counters.

\paragraph{Consequence.}
The effect needs the largest checkpoint, batch 64, fp32 and this inner dimension
together: batch~1 gives the expected ratio of $714/648$, the cached rows stay at
$1.02$, and the smaller checkpoints lie away from the range. We report the row as
measured, since it is the time such a call takes on this device; TF32 removes the
effect but changes the numerics. The fitted slope is unaffected ($0.811$ against
$0.812$\,ms per step), so the scaling in Sec.~\ref{sec:streaming} stands, but
contexts 11328 to 20736 are pessimistic by $9$ to $25\%$ and should not carry
point comparisons.

\newpage
\section{A window is a third mechanism, and a different forecast}
\label{app:windowing}

Both mechanisms of Sec.~\ref{sec:model-comparison} condition the forecast on the
entire observed sequence, and both reproduce the forecast of a full pass exactly:
the KV cache retains every past chunk and the recurrent model keeps a compressed
memory that each update adapts in place. What they change is the latency and
memory of one update, not the context the forecast depends on. A fixed window
bounds both for any architecture, causal or not, but it shortens that context.

\begin{figure}[ht]
\centering
\includegraphics{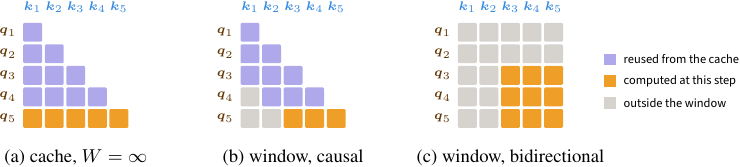}
\caption{A window bounds latency and memory by shrinking what the forecast
reads, drawn at $W=3$; cells are the attention scores $\Bq_i^{\top}\Bk_j$ of query chunk $i$
against key chunk $j$. \textbf{(a)}~the cache keeps every past chunk.
\textbf{(b)}~a window is still causal, so the arriving chunk appends a row and
the cache holds $W$ entries. \textbf{(c)}~on a bidirectional stack sliding the
window revises every representation inside it. One update takes
$\mathcal{O}(n)$, $\mathcal{O}(W)$ and $\mathcal{O}(W^{2})$ in time.}
\label{fig:window-mechanism}
\end{figure}

\begin{table}[h]
\centering
\caption{One refresh under a fixed truncation window, batch size 1, at the released context
limit of each architecture and at a common window of $2048$ steps, the smallest
limit in the set. Every update re-encodes the last $W$ observed steps; no row reuses a
cache or a state, and no row uses windowed attention inside a layer. Rows time one
deterministic forward pass, except Toto-1.0, which draws its released number of sampled
paths per refresh.}
\label{tab:window-edge}
\begin{threeparttable}
\small
\setlength{\tabcolsep}{5pt}
\begin{tabular}{l r rrr r}
\toprule
& & \multicolumn{3}{c}{latency per refresh, ms} & Pi 5 \\
\cmidrule(lr){3-5}
Architecture & $W$ & Pi 5 & C6015 & C6025 & GB \\
\midrule
\multicolumn{6}{l}{\itshape Causal, recurrent} \\
TiRex-2           & 8192  & $823 \pm 14$      & $2\,130 \pm 18$     & $800 \pm 5$        & $1.112 \pm 0.006$ \\
                  & 2048  & $282 \pm 4$       & $695 \pm 35$        & $268 \pm 14$       & $1.108 \pm 0.006$ \\
\addlinespace[2pt]
TiRex             & 2048  & $718 \pm 23$      & $972 \pm 7$         & $434 \pm 4$        & $0.900 \pm 0.003$ \\
\addlinespace
\multicolumn{6}{l}{\itshape Causal, attention} \\
Toto-1.0\tnote{a} & 4096  & $77\,842 \pm 332$ & $205\,842 \pm 971$  & $55\,954 \pm 80$   & $1.693 \pm 0.006$ \\
                  & 2048  & $41\,713 \pm 101$ & $104\,977 \pm 386$  & $29\,248 \pm 5309$ & $1.580 \pm 0.002$ \\
\addlinespace[2pt]
Toto-2.0-22m      & 4096  & $177 \pm 2$       & $422 \pm 5$         & $121 \pm 9$        & $0.872 \pm 0.001$ \\
                  & 2048  & $141 \pm 1$       & $313 \pm 4$         & $100 \pm 1$        & $0.865 \pm 0.001$ \\
\addlinespace[2pt]
TimesFM-2.5       & 16256 & $8\,075 \pm 41$   & $20\,906 \pm 38$    & $4\,710 \pm 32$    & $1.864 \pm 0.004$ \\
                  & 2048  & $1\,440 \pm 7$    & $2\,830 \pm 7$      & $886 \pm 14$       & $1.699 \pm 0.003$ \\
\addlinespace
\multicolumn{6}{l}{\itshape Non-causal} \\
Chronos-2         & 8192  & $17\,896 \pm 101$ & $6\,103 \pm 4$      & $1\,823 \pm 5$     & $1.231 \pm 0.001$ \\
                  & 2048  & $4\,706 \pm 63$   & $1\,613 \pm 4$      & $478 \pm 12$       & $1.207 \pm 0.001$ \\
\addlinespace[2pt]
Chronos-2-small   & 8192  & $3\,184 \pm 111$  & $1\,336 \pm 2$      & $328 \pm 2$        & $0.873 \pm 0.001$ \\
                  & 2048  & $831 \pm 9$       & $403 \pm 1$         & $96 \pm 1$         & $0.861 \pm 0.001$ \\
\addlinespace[2pt]
Chronos-Bolt-base & 2048  & $658 \pm 10$      & $1\,195 \pm 21$     & $259 \pm 1$        & $1.538 \pm 0.007$ \\
\bottomrule
\end{tabular}
\begin{tablenotes}[flushleft]
\footnotesize
\item[a] Toto-1.0 is the only configuration with a sampling head and is timed here at the
released default of $256$ sampled paths per refresh, whereas Fig.~\ref{fig:refresh-latency} and Tab.~\ref{tab:energy-edge} time a
single path. Its rows are therefore not comparable with the remaining eight;
dividing by $256$ does not recover a single-path cost, since the paths share the fixed
cost of one call.
\end{tablenotes}
\end{threeparttable}
\end{table}
\paragraph{The interval between arrivals sets the refresh budget, and it does not grow with the history.} A chunk arrives every $p$ steps and the next forecast is due before the following chunk arrives. Everything it conditions on except the arriving chunk is available earlier, so a system holding its context in a retained state has only the new chunk left once the deadline starts running. The interval is constant while the history grows, so an update whose latency grows with the context leaves the interval at some stream length, set by the device rather than by the model. A window is bounded but re-encodes $W$ chunks at every arrival instead of one, using the interval to rebuild representations it held one step earlier; Table~\ref{tab:window-edge} gives that re-encode as 1.3 to 18 times the native-minimum refresh at
$W = 2048$ on the Raspberry Pi 5, across the eight deterministic configurations.

\paragraph{A window bounds latency and memory for a causal architecture.}
Restricting attention to the last $W$ chunks preserves the property
Sec.~\ref{sec:model-comparison} relies on, that new observations do not change
earlier representations: the arriving chunk appends a row to the score grid and
the cache becomes a ring buffer that evicts its oldest entry
(Fig.~\ref{fig:window-mechanism}b). One update then takes $\mathcal{O}(W)$ in
time and $\mathcal{O}(W)$ in memory instead of $\mathcal{O}(n)$, and every
causal architecture in the set admits this with no change beyond the cache we
already build. A bidirectional encoder still re-encodes the window at every
update, $\mathcal{O}(W^{2})$ in time, which bounds the floor
Fig.~\ref{fig:refresh-latency} measures without producing a state update.
Table~\ref{tab:window-edge} measures that re-encode for every architecture on the three edge devices: at
$W = 2048$ on the Raspberry Pi 5 it takes 1.3 to 18 times the native-minimum refresh
of the same model across the eight deterministic configurations, and more at each
released context limit.

\paragraph{A window is bounded, but it does not overtake a retained state.}
Because a truncation window of $W$ steps conditions the forecast on exactly the last $W$ observations, the columns of
Table~\ref{tab:acc-agg} give the accuracy of the windowed baseline directly and Table~\ref{tab:window-edge} its latency
on the edge devices. Paired at the best measured window of each architecture, no configuration is both faster and more accurate
than the incremental refresh of TiRex-2, which scores $45.3$ on the untruncated series at about $120$\,ms on the
Raspberry~Pi~5. Chronos-2 is the only architecture more accurate at all, $47.6$ at $W = 2048$, and pays $4\,706$\,ms per
refresh, $39$ times the incremental one; Chronos-2-small reaches the same $45.3$ at $831$\,ms, seven times. The
remaining five are behind on both axes: TiRex at $37.0$ and $718$\,ms, Toto-2.0-22m at $36.9$ and $177$\,ms, Toto-1.0 at
$35.0$ and $77\,842$\,ms for its $256$-path refresh (Tab.~\ref{tab:window-edge}, note~a), TimesFM-2.5 at $34.8$ and
$8\,075$\,ms, and Chronos-Bolt at $34.8$ and $658$\,ms. Peak memory
does not separate them either, as every windowed row lies between $0.86$ and $1.86$\,GB against $1.2$\,GB for the
incremental refresh. The conclusion of Sec.~\ref{sec:conclusion} therefore does not rest on the exactness requirement:
dropping it and granting every architecture its best measured window leaves the ordering unchanged.

\paragraph{A window changes what the forecast conditions on.}
Under truncation the forecast reads the last $W$ chunks; under windowed attention
each of the $L$ windowed layers widens the dependency by $W-1$ chunks, so the
forecast after chunk $n$ is a function of the last $R = L\,(W-1)+1$ chunks,
$\hat{x}_{n+1} = f(x_{n-R+1:\,n})$, rather than of $x_{1:n}$. The two are equal
only if nothing older than $R$ chunks affects the prediction. A windowed architecture therefore fails the exactness requirement that every
architecture we report satisfies.

\newpage
\section{Equivalence of incremental and full recomputation}
\label{app:equivalence}

For every causal configuration we ingest a context of $2048$ steps incrementally and compare the predictive mean
with that of a full forward pass over the same context (\Cref{eq:correctness}), at batch size $1$ in float32 with
TF32 disabled. All configurations pass \texttt{allclose} with $\mathrm{atol}=10^{-5}$ and $\mathrm{rtol}=10^{-4}$
(\Cref{tab:equivalence}). The relative error of Toto-1.0 is larger because its predictive mean is small ($0.021$,
against $0.13$ to $9.0$ for the other models). Its absolute error is in the same range as the others.

\begin{table}[h]
\centering
\small
\caption{Deviation of the incremental forecast from full recomputation.}
\label{tab:equivalence}
\begin{tabular}{@{}lrr@{}}
\toprule
Architecture & max.\ abs.\ error & max.\ rel.\ error \\
\midrule
TiRex-2       & $3.73\times10^{-8}$ & $3.29\times10^{-7}$ \\
TiRex         & $4.77\times10^{-7}$ & $3.12\times10^{-7}$ \\
Toto-1.0      & $4.75\times10^{-7}$ & $2.23\times10^{-5}$ \\
Toto-2.0-22m  & $3.58\times10^{-7}$ & $3.26\times10^{-7}$ \\
Toto-2.0-2.5B & $3.58\times10^{-7}$ & $3.28\times10^{-7}$ \\
TimesFM-2.5   & $9.54\times10^{-7}$ & $1.06\times10^{-7}$ \\
\bottomrule
\end{tabular}
\end{table}

\section{Software stack}
\label{app:softwarestack}

All configurations run through their released packages (tirex 1.0.0, toto-ts 0.1.3, toto-2 2.0.0, timesfm 2.0.2,
chronos-forecasting 2.2.2) under PyTorch 2.9.1 built against CUDA 12.8 and cuDNN 9.8.0, on driver 590.48.01, in
float32 except Chronos-Bolt (bfloat16, as released). TF32 is disabled for matrix multiplication \newline
(\texttt{float32\_matmul\_precision} is \texttt{highest}); cuDNN's TF32 path is left at its default, which the
measured models do not reach. Scaled dot-product attention is left to the PyTorch dispatcher with all four
backends available. On the A40 host, BLAS and LAPACK are Intel MKL 2024.2, and both the intra- and inter-op thread
pools hold 64 threads. A reported latency is the median of 64 synchronised calls after warmup at the same context.
Priming the state is not timed. Memory is the allocated peak of that window.

\end{document}